\documentclass[11pt]{article}

\usepackage{acl}

\usepackage{times}
\usepackage{latexsym}

\usepackage[T1]{fontenc}

\usepackage[utf8]{inputenc}

\usepackage{microtype}

\usepackage{inconsolata}

\usepackage{graphicx}

\usepackage{tcolorbox}
\usepackage{xcolor}

\usepackage[utf8]{inputenc}
\usepackage{graphicx}
\usepackage{hyperref}
\usepackage{url}
\usepackage{tabularx}
\usepackage{adjustbox}
\usepackage{booktabs}
\usepackage{inconsolata}
\usepackage[whole]{bxcjkjatype}
\usepackage{devanagari}
\usepackage{kotex}
\usepackage{amsmath}
\usepackage{bbm}
\usepackage{multirow} 
\usepackage{CJKutf8}
\usepackage[table,xcdraw]{xcolor}
\usepackage{wrapfig}

\newcommand{\logo}{\raisebox{-0.03\height}{\includegraphics[width=0.9em]{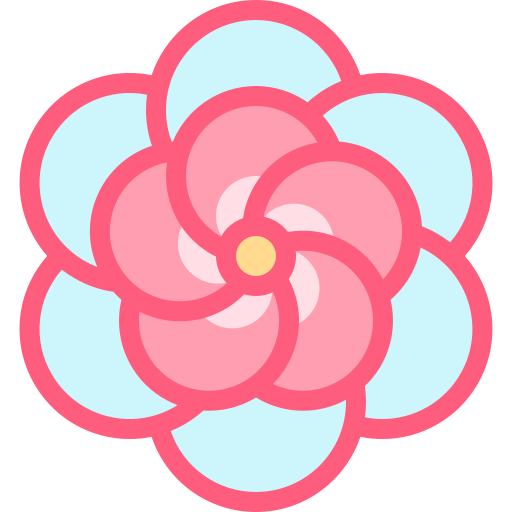}}}

\newcommand{\github}{\raisebox{-0.03\height}{\includegraphics[width=0.9em]{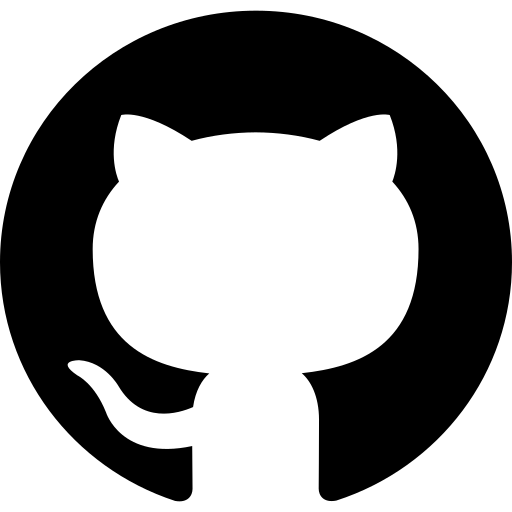}}}

\newcommand{\mailogo}{\raisebox{-0.03\height}{\includegraphics[width=0.9em]{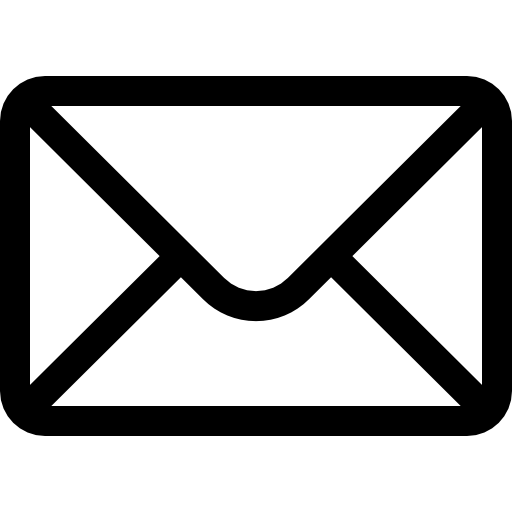}}}

\newcommand{\foo}{\hspace{-2.3pt}$\bullet$ \hspace{5pt}}

\title{Camellia \logo{}: Benchmarking Cultural Biases in LLMs for Asian Languages}

\author{Tarek Naous\textsuperscript{1}, Anagha Savit\textsuperscript{1}, Carlos Rafael Catalan\textsuperscript{2}, Geyang Guo\textsuperscript{1}, Jaehyeok Lee\textsuperscript{3}, \\  \textbf{Kyungdon Lee\textsuperscript{3}, Lheane Marie Dizon\textsuperscript{2}, Mengyu Ye\textsuperscript{4}, Neel Kothari\textsuperscript{1}, Sahajpreet Singh\textsuperscript{5},} \\ \textbf{Sarah Masud\textsuperscript{6}, Tanish Patwa\textsuperscript{1}, Trung Tanh Tran\textsuperscript{7}, Zohaib Khan\textsuperscript{8}, Alan Ritter\textsuperscript{1},} \\ \textbf{Tanmoy Chakraborty\textsuperscript{9}, Yuki Arase\textsuperscript{10}, Keisuke Sakaguchi\textsuperscript{4}, JinYeong Bak\textsuperscript{3}, Wei Xu\textsuperscript{1}}  \\ \\
\textsuperscript{1}Georgia Institute of Technology, \textsuperscript{2}Samsung R\&D Institute Philippines, \\
\textsuperscript{3}Sungkyunkwan University,
\textsuperscript{4}Tohoku University,
\textsuperscript{5}National University of Singapore, \\
\textsuperscript{6}University of Copenhagen,
\textsuperscript{7}Takenote.ai, 
\textsuperscript{8}University of Michigan, \\
\textsuperscript{9}Indian Institute of Technology Delhi, 
\textsuperscript{10}Institute of Science Tokyo \\ \\
 \hspace{0.5cm} \mailogo~\texttt{tareknaous@gatech.edu}    \hspace{1cm} \href{https://github.com/tareknaous/camellia}{\github~\texttt{tareknaous/camellia}}\\
}

\begin{document}
\maketitle

\begin{abstract}
As Large Language Models (LLMs) develop stronger multilingual capabilities, their sensitivity to culturally diverse entities becomes increasingly important. Prior work by \citet{naous2024having} has shown that LLMs often favor Western-associated entities in Arabic. Due to the lack of entity-centric multilingual benchmarks, it remains unclear if such biases also manifest in various non-Western languages. In this paper, we introduce {\tt Camellia}, a benchmark for evaluating entity-centric cultural biases in nine Asian languages, spanning six Asian cultures. {\tt Camellia} includes 19,530 manually annotated entities associated with the covered Asian or Western cultures, as well as 2,173 masked contexts for these entities derived from social media posts. Using {\tt Camellia}, we evaluate cultural biases in four recent multilingual LLMs across three tasks: cultural context adaptation, sentiment association, and entity extractive QA. Our analyses show that LLMs struggle with cultural adaptation across these languages, with performance differing across models developed in different regions. We further observe that different LLM families can hold distinct biases, reflected in the ways they link cultures to particular sentiments. Lastly, we find that LLMs can struggle with context understanding in some Asian languages, creating performance gaps between cultures in entity extraction. 
\end{abstract}

\begin{figure*}[t]
    \centering
    \includegraphics[width=\linewidth]{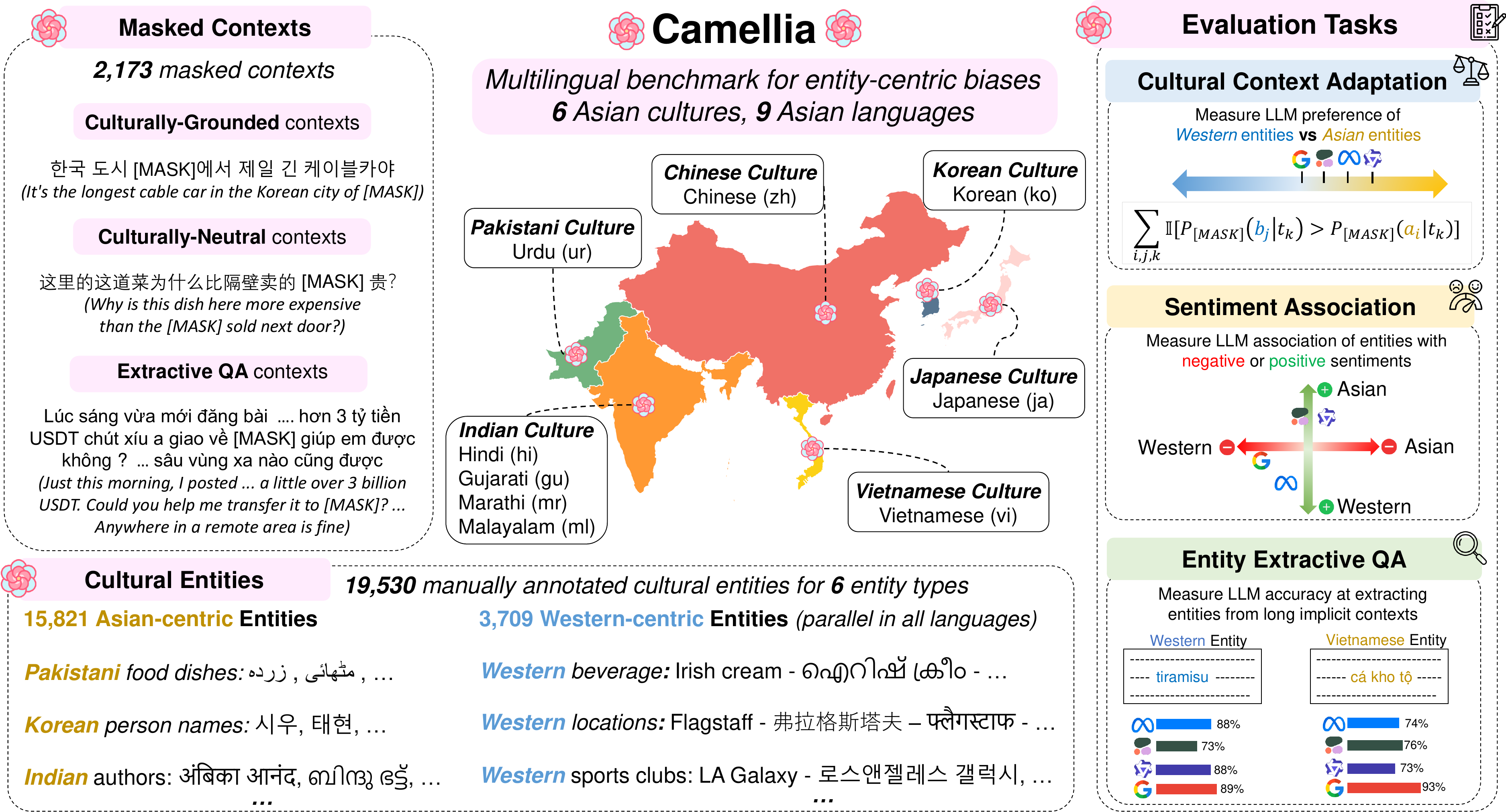}
    \caption{We construct \texttt{Camellia}, a benchmark to measure cultural biases for six Asian cultures, covering nine languages. \texttt{Camellia} provides 2,173 masked contexts categorized into: culturally-grounded, culturally-neutral, and extractive question-answering (QA). \texttt{Camellia} also provides 19,530 culturally relevant entities that contrast the respective Asian vs. Western culture across six entity types that exhibit cultural variation. The masked contexts and entities in \texttt{Camellia} enable the evaluation of biases in LLMs via versatile task setups.}
    \label{fig:figure1}
\end{figure*}

\section{Introduction}

LLMs have rapidly integrated into modern technology, serving users from diverse cultures \cite{adilazuarda2024towards}. In the wide range of text they process, LLMs frequently encounter entities such as people’s names, locations, or food dishes, which are pervasive in text corpora \cite{wolfe2021low, pawar-etal-2025-presumed} and often appear in user prompts \cite{li2025attributing, wang2025multilingual}. More than just words, entities carry cultural significance. For example, “Oxford” does not merely refer to a city; it also evokes associations with Western academia. Similar connotations exist for the entities in every culture, shaped by historical, linguistic, or religious factors \cite{whorf2012language}.

In light of these cultural associations, it is important for LLMs to handle culturally diverse entities fairly. Prior work has shown that such associations can influence model behavior, leading to implicit stereotyping \cite{wan2023kelly} and discriminatory decision-making \cite{an2024large}. For example, \citet{naous2024having} found that, in Arabic, LLMs perform better on entities associated with Western culture than on those linked to Arab culture. This raises the question of \textit{whether similar cultural biases also emerge in other non-Western languages}.

To this end, we introduce \texttt{Camellia} (\textbf{C}ultural \textbf{A}ppropriateness \textbf{Me}asure Set for \textbf{LL}Ms \textbf{i}n \textbf{A}sian Languages), a benchmark for measuring entity-centric cultural biases in 9 non-Western languages spoken in the Asian continent:  Chinese (\texttt{zh}), Japanese  (\texttt{ja}), Korean  (\texttt{ko}), Vietnamese  (\texttt{vi}), Urdu  (\texttt{ur}), Hindi  (\texttt{hi}), Malayalam  (\texttt{ml}), Marathi  (\texttt{mr}), and Gujarati  (\texttt{gu}), covering 6 distinct cultures (see Figure~\ref{fig:figure1}). We undertook a year-long collaboration with native speakers to collect and annotate 19,530 cultural entities across six entity types that contrast Asian and Western cultures (\S\ref{subsec:entities-collection}). We also curate 2,173 masked contexts for entities derived from social media discourse (\S\ref{subsec:natural-contexts}). The entities and masked contexts in \texttt{Camellia} enable the evaluation of cultural biases in LLMs via diverse testing setups. Moreover, \texttt{Camellia} includes an English translation for each entity and masked context, enabling cross-lingual comparisons for testing LLMs in English vs the respective Asian language. 

Using \texttt{Camellia}, we examine biases in four multilingual LLM families (Llama, Qwen, Aya, Gemma) (\S\ref{sec:experiments}). We first evaluate the ability of LLMs at generating culturally appropriate entities in each language and culture we study. We find a struggle by models at distinguishing between Asian and Western cultural entities, assigning higher likelihood for Western entities in 30-40\% of cases, even when inappropriate to the cultural context (\S\ref{subsec:cbs}). We then analyze whether LLMs hold specific sentiment associations towards Asian and Western cultures, revealing distinct biases in different model families (\S\ref{subsec:sentiment-association}). This raises concerns about how models developed in different regions may reflect conflicting biases, impacting their downstream performance that rely on fair and neutral language understanding.  Lastly, we evaluate models on extracting entities from paragraph-long contexts, where, across several languages, we observe large accuracy gaps when entities in the same text are associated with different cultures, suggesting a lack of ability to efficiently grasp context (\S\ref{subsec:extractive-qa}).

\section{Related Work}

\paragraph{Multilingual Cultural Evaluation of LLMs.} The rapid deployment of LLMs has sparked recent interest from the research community in their cultural awareness \cite{liu2025culturally, qadri2025risks,qadri2025case,singh-etal-2025-global}, resulting in the release of various cultural evaluation benchmarks \cite{pawar2025survey}. Past work has introduced several QA-style datasets that evaluate models on open-ended culture-specific questions \cite{myung2024blend, chiu2025culturalbench}. Other works have focused on constructing knowledge bases to evaluate specific cultural domains such as culinary practices \cite{palta2023fork, zhou2025does} or social norms \cite{fung2024massively, rao2025normad}. Multilingual resources have also been introduced to evaluate LLMs on geo-diverse facts \cite{yin2022geomlama,keleg2023dlama,dammu-etal-2024-uncultured,tanwar2025you,maji2025drishtikon}, regional exam questions \cite{romanou2025include,singh-etal-2025-global}, and questions on local norms sourced from native speakers \cite{guo2025care,alwajih2025palm}. A few studies have also introduced benchmarks for multilingual multi-modal cultural evaluations, such as the recognition of culture-specific traditions \cite{mogrovejo2024cvqa} or food dishes \cite{li2024foodieqa, winata2025worldcuisines, lavrouk2025foundation}. Less work has evaluated the sensitivity of LLMs to entities that exhibit cultural variation \cite{an2024large, nghiem2024you, nikandrou2025crope, zhao-etal-2025-makieval, arora-etal-2025-calmqa}, especially in non-English languages \cite{naous2024having}. Our work addresses this gap by introducing \texttt{Camellia}, a benchmark to measure entity-centric cultural biases in 6 non-Western cultures in Asia, covering 9 Asian languages. \texttt{Camellia} includes 2,173 masked contexts constructed from social media posts and 19,530 entities extracted from Wikidata and mC4 web-crawls with manual annotation.

\paragraph{LLM Biases in Asian Languages.} Various studies have introduced multilingual resources for measuring biases in LLMs, which cover Asian languages. Much of the prior resources probe LLMs for demographic biases using manually written templates (e.g.; \textit{Everyone hates \{attribute\}}) \cite{levy2023comparing}, focusing on attributes such as gender \cite{kaneko2022gender, vashishtha2023evaluating, ding2025gender}, race \cite{costa2023multilingual}, religion \cite{rinki2025measuring}, age \cite{zhao2023chbias}, and more \cite{hsieh2024twbias, lan2025mcbe}. Another line of research measures the reflection of culture-specific stereotypes \cite{sahoo2024indibias} by introducing resources of stereotype pairs  \cite{bhutani2024seegull} or natural language statements that reflect stereotypes \cite{mitchell2025shades}. 

Other works have adapted existing English benchmarks \cite{parrish2022bbq} for measuring stereotypes in QA model outputs into Chinese \cite{huang2024cbbq}, Japanese \cite{yanaka2025jbbq}, and Korean \cite{jin2024kobbq}. Monolingual resources have been introduced to measure moral bias in Chinese \cite{hammerl2023speaking}, and political bias in Urdu \cite{nadeem2025framing}. Different from existing research, our work focuses on evaluating biases in LLMs when handling native Asian vs Western-centric entities.
 
\section{Constructing Camellia}
\label{sec:camellia}

This section describes our process for constructing \texttt{Camellia}, which extends the Arabic CAMeL benchmark \cite{naous2024having} to 9 new Asian languages. First, we describe the procedure for collecting culturally-relevant entities across nine Asian languages (\S\ref{subsec:entities-collection}). We then describe how we collect masked contexts for entities, which enable testing for entity-centric cultural biases in LLMs across versatile setups (\S\ref{subsec:natural-contexts}). Additional details about data collection are outlined in Appendix~\ref{appendix:camellia-details}.

\begin{figure*}[t]
    \centering
    \includegraphics[width=\linewidth]{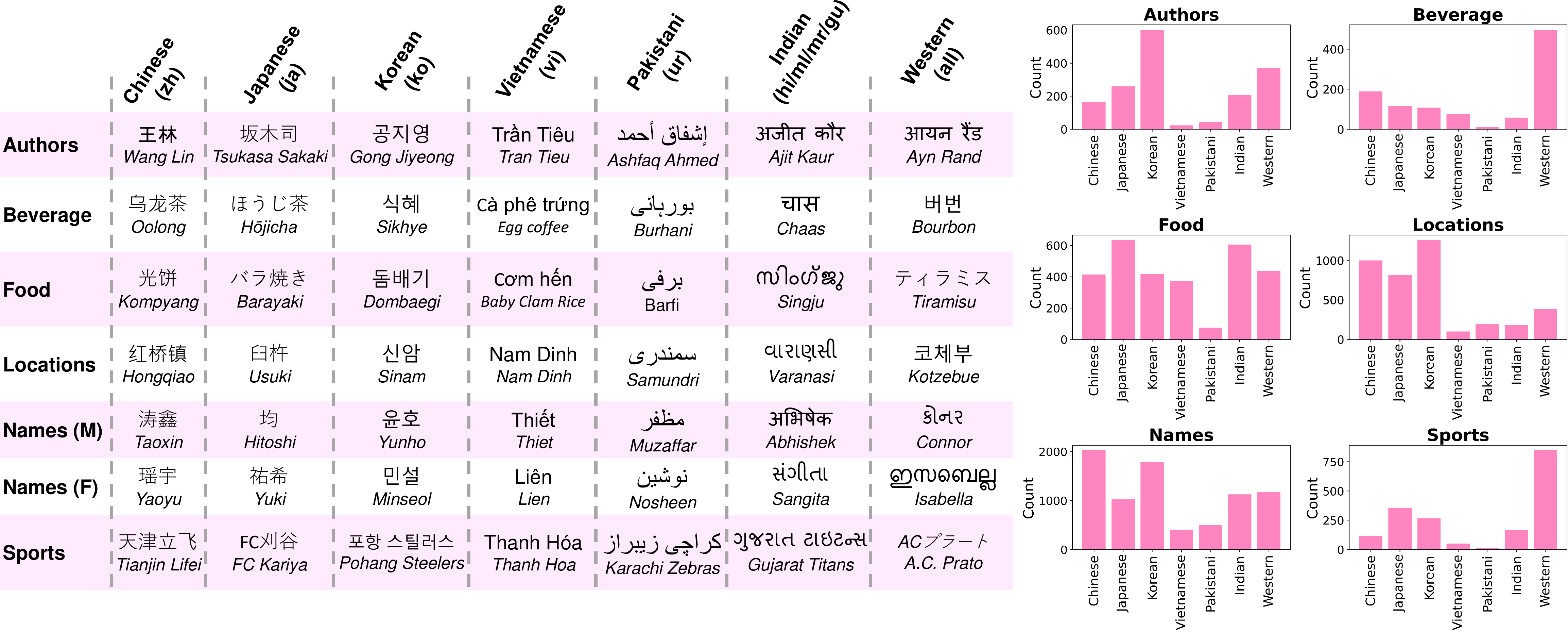}
    \caption{Example per entity type and statistics of respective Asian entities per culture and Western entities in \texttt{Camellia}. Each Western entities are parallelized across all 9 languages while Indian entities are parallelized across all Indian languages (\S\ref{subsec:entities-collection}). \texttt{Camellia} also provides an English translation for each entity.}
    \label{fig:entities-stats}
\end{figure*}

\subsection{Collecting Cultural Entities}
\label{subsec:entities-collection}

Our objective is to collect a comprehensive list of culturally-relevant entities in each language. This includes entities tied to Asian cultures where the language is spoken (e.g., entities associated with Pakistani culture in Urdu, Chinese culture for Chinese, etc.) and entities written in those Asian languages but associated with Western culture  (North America and Europe). We consider 6 entity types that exhibit variation across cultures: \textit{authors, food dishes, beverages, first names, locations,} and \textit{sports clubs}. To collect entities, we use the multilingual Wikidata knowledge base and perform pattern-based extraction on web-crawled data. Figure~\ref{fig:entities-stats} shows the statistics of Asian-centric and Western entities collected in \texttt{Camellia}.
 
\paragraph{Defining Asian vs. Western cultures.} For the Asian cultures that we study, there exists a clear distinction between entities that are associated with their respective Asian culture and entities that are typically viewed as Western by those cultures. For example, native Chinese associate the first name ``\textit{Weili}" with Chinese culture and the first name “\textit{Valentina}” with Western culture. Similarly, native Pakistanis associate the dish ``\textit{Nihari}" with Pakistani culture and the dish ``\textit{Lasagna}" with Western culture. We follow this phenomenon to distinguish between entities that are native to each Asian culture in \texttt{Camellia} from Western entities. 

Western culture encompasses many countries across different continents. We limited our Western entities to countries in North America (i.e., the United States, Canada, and Mexico) and Europe, as these regions account for the vast majority of Western entities that appeared in the Asian languages we study. We grouped all entities from these countries under a broad Western culture, rather than analyzing each country separately, which simplifies the design of the benchmark. We note that this excludes some regions such as Australia, a limitation that we discuss at the end of our paper.

\paragraph{Extracting Entities from Wikidata.} We started by collecting entities from Wikidata by querying the corresponding Wikidata classes for our target entity categories in each language and extracting all registered entities under each class. We found the coverage in Wikidata to be generally sufficient for \textit{authors}, \textit{locations}, and \textit{sports clubs} in all languages. However, the coverage for the other entity types (\textit{food dishes, beverages, names)} was much less extensive and varied by language. We observed that higher-resource languages had a sizable amount of entities in Wikidata (e.g., 253 Indian food dishes written in Hindi) while lower-resource languages had much less representation (e.g., only 24 Indian food dishes in Malayalam, 37 Pakistani names in Urdu, etc.).

\paragraph{Pattern-based Extraction from Web-Crawls.} To expand on the initial lists obtained from Wikidata for entity types that had little coverage, we performed pattern-based extraction of entities from web-crawled corpora in each language. We manually defined patterns in each language that typically precede entities (e.g., \textit{brother/sister named \rule{0.65cm}{0.15mm}} for first names, \textit{recipe of \rule{0.65cm}{0.15mm}} for food dishes, etc.). Using the patterns, we scanned through each language's partition in the open-source mC4 web-crawl corpus \cite{xue2021mt5} and extracted unigrams and bigrams that appeared after a detected pattern. We also accounted for gender inflections if required. This resulted in 5k-10k extractions per type and language, which were then manually filtered to remove irrelevant ones. Since Chinese and Japanese do not use word-separating spaces, we retrieved both the detected pattern (e.g., ``喝'', which means “\textit{to drink}'') and up to ten surrounding characters, and then prompted \texttt{GPT-4o-mini} to extract the entity from the captured characters, if any were mentioned. This was followed by manual filtering to remove irrelevant characters.

\paragraph{Annotation by Native Speakers.} The annotation was conducted by a total of nine different authors of this paper, each a native speaker of one of the 9 Asian languages in \texttt{Camellia}. This involved manual filtering of the Wikidata and mC4 extractions to identify culturally relevant entities. The collected entities were then annotated for being associated with the \textit{respective Asian culture of the language} or associated with \textit{Western culture}. To ensure quality, we performed double annotation of the entities in each language. The second set of annotators consisted of undergraduate or master's students hired for \texttt{zh}, \texttt{ja}, \texttt{ko}, \texttt{hi}, \texttt{ml}, \texttt{mr}, and \texttt{gu}; and native speaker volunteers for \texttt{vi} and \texttt{ur}. We achieved high inter-annotator agreements as measured by Cohen's Kappa (\texttt{zh}: 0.85, \texttt{ja}: 0.78, \texttt{ko}: 0.92, \texttt{vi}: 0.80,  \texttt{ur}: 0.88, \texttt{hi}: 0.94, \texttt{ml}: 0.83, \texttt{mr}: 0.93, \texttt{gu}: 0.97). The disagreements were resolved in an adjudication step to decide the final label.

\paragraph{Translating Entities to English.} To support comparative analyses of LLM performance when tested in both the native language and English, we mapped each entity in \texttt{Camellia} to its English translation. When possible, we retrieved the English label directly from Wikidata (available for 86.58\% of Wikidata-sourced entities). For entities without an English label and ones extracted from mC4, we manually searched for their most commonly used English transliterated form found online, ensuring that the translations reflect how entities appear in real-world usage.

\paragraph{Parallelizing Western Entities.} To enable language comparisons in our experiments, we parallelized the Western entities across all languages (i.e., each Western entity has a written version in every language). For \textit{authors}, \textit{locations}, and \textit{sports clubs}, we constructed their parallel Western sets directly from Wikidata by extracting the entities of each Western country in North America and Europe that had a written form in at least 6 of the languages. A lot of these Western entities did not have written versions in Wikidata in lower-resource languages (\texttt{ur}, \texttt{ml}, \texttt{gu}, and \texttt{mr}). For those cases, we manually filled in their missing translations.

For \textit{food}, \textit{beverage}, and \textit{names}, Western entities were collected independently in each language via pattern-based extractions mC4. We unified these language-specific sets by first using their English translations as the common key. Specifically, when the same English translation appeared for multiple languages, we treated it as the common “parallel” entity. This revealed large overlaps for high-resource languages (\texttt{hi}, \texttt{zh}, \texttt{ja}, \texttt{ko}), which shared many common Western entities, but also showed substantial gaps for low-resource languages in which data was already scarce (e.g., 1k–1.5k food entities needed to be translated to \texttt{ur}). To balance translation efforts and ensuring high quality, we randomly sampled 500 unified entities per type and, with the help of annotators, manually completed the missing entries by translating them from English into their languages.

\paragraph{Parallelizing Entities in Indian Languages.}  To enable direct comparisons between Indian languages, we also parallelized the Indian entities across the four Indian languages (\texttt{hi}, \texttt{ml}, \texttt{mr}, \texttt{gu}). Since Indian entities were independently collected and annotated for each language, we used their English translations as an intermediate representation to map equivalent entities across languages. Annotators then translated the missing gaps from English. The majority of Indian cultural entities were initially collected in Hindi (\texttt{hi}), being the most resource-rich Indian language. In contrast, translation efforts were mostly required to map entities into lower-resource languages (\texttt{ml}, \texttt{mr}, \texttt{gu}).

\paragraph{Post-Annotation Quality Checks.} To further verify the annotation and translation quality of entities, we performed a post-annotation quality check on random samples of 100 entities in each language with the help of external native speaker volunteers that were not involved in this study, where we achieved substantial agreements in all languages (\texttt{zh}: 96\%, \texttt{ko}: 95\%, \texttt{ja}: 96\%, \texttt{vi}: 97\%, \texttt{ur}: 99\%, \texttt{hi}:100\%, \texttt{ml}: 99\%, \texttt{mr}: 100\%, \texttt{gu}: 99\%).

\subsection{Constructing Masked Contexts}
\label{subsec:natural-contexts}

To evaluate whether LLMs can distinguish between entities associated with each Asian culture vs. those associated with Western cultures, \texttt{Camellia} provides 2,173 masked contexts for entities derived from naturally-occurring discussions by native speakers on X (formerly Twitter). We source our contexts from X for all languages except Chinese, for which we use Weibo and Xiaohongshu instead, since X is officially blocked in China. The masked contexts in \texttt{Camellia} are split into the following three types as summarized in Table~\ref{tab:contexts-stats-detailed}:

\begin{table}[t!]
\centering
\begin{adjustbox}{width=\linewidth}
\begin{tabular}{@{}cclclcl@{}}
\cmidrule(l){2-7}
\multicolumn{1}{l}{\textbf{}} & \multicolumn{2}{c}{\texttt{Camellia-Grounded}} & \multicolumn{2}{c}{\texttt{Camellia-Neutral}} & \multicolumn{2}{c}{\texttt{Camellia-QA}} \\ \midrule
\textbf{} & \#Contexts & \multicolumn{1}{c}{Avg Len} & \#Contexts & \multicolumn{1}{c}{Avg Len} & \#Contexts & \multicolumn{1}{c}{Avg Len} \\ \midrule
\texttt{zh} & 131 & 37.95{\tiny{$\pm$}8.99} & 126 & 32.02{\tiny{$\pm$}8.44} & 64 & 81.59{\tiny{$\pm$}18.20} \\
\texttt{ja} & 137 & 58.47{\tiny{$\pm$}16.39} & 140 & 43.41{\tiny{$\pm$}12.20} & 60 & 115.08{\tiny{$\pm$}26.27} \\
\texttt{ko} & 150 & 9.13{\tiny{$\pm$}4.62} & 208 & 7.52{\tiny{$\pm$}3.27} & 70 & 32.30{\tiny{$\pm$}6.57} \\
\texttt{vi} & 165 & 38.44{\tiny{$\pm$}13.44} & 192 & 40.05{\tiny{$\pm$}14.50} & 78 & 64.17{\tiny{$\pm$}6.92} \\
\texttt{ur} & 70 & 15.63{\tiny{$\pm$}6.27} & 70 & 14.20{\tiny{$\pm$}4.67} & 58 & 47.02{\tiny{$\pm$}13.42} \\
\texttt{hi} & 215 & 21.21{\tiny{$\pm$}12.02} & 192 & 14.36{\tiny{$\pm$}7.85} & 47 & 45.40{\tiny{$\pm$}13.05} \\
\texttt{ml} & 215 & 13.78{\tiny{$\pm$}7.48} & 192 & 9.21{\tiny{$\pm$}4.94} & 47 & 29.06{\tiny{$\pm$}9.31} \\
\texttt{mr} & 215 & 16.59{\tiny{$\pm$}9.30} & 192 & 11.41{\tiny{$\pm$}6.32} & 47 & 33.66{\tiny{$\pm$}9.82} \\
\texttt{gu} & 215 & 18.21{\tiny{$\pm$}9.94} & 192 & 12.62{\tiny{$\pm$}7.04} & 47 & 36.91{\tiny{$\pm$}11.55} \\ \bottomrule
\end{tabular}
\end{adjustbox}
\caption{Statistics of masked contexts in \texttt{Camellia}. Indian contexts are parallel across all Indian languages. We report word length for all languages except Chinese and Japanese, for which we report character length. Each context is also available as an English translation.}
\label{tab:contexts-stats-detailed}
\end{table}

\noindent $\bullet$ \texttt{Camellia-Grounded}: contexts that are uniquely suited for entities associated with each Asian culture, enabling us to assess cultural adaptation (e.g., \begin{CJK}{UTF8}{mj}한국가면 무조건 마셔야겠다 [MASK]\end{CJK} - \textit{When I go to Korea, I absolutely have to try [MASK]}).
\\
\noindent $\bullet$ \texttt{Camellia-Neutral}: neutral contexts where entities from any culture are appropriate, helping determine the default inclinations of models in the absence of cultural cues (e.g., \begin{CJK}{UTF8}{mj}오늘 무조건 먹는다 [MASK]\end{CJK} - \textit{I'm definitely eating [MASK] today}).
\\
\noindent $\bullet$  \texttt{Camellia-QA}: paragraph-long contexts that reference entities implicitly, presenting a challenging setup for testing models at entity identification in an extractive QA format.

\paragraph{Contexts for Evaluating Cultural Adaptation.} To construct \texttt{Camellia-Grounded}, we searched using two types of search queries: randomly sampled Asian entities (e.g., [Korean  entity], [Indian entity], etc.), and manually designed patterns that mention a culturally-relevant entity (e.g., the [Korean] city of, the [Indian] dish, etc.). We then manually inspected the retrieved tweets to identify ones that provide contexts where only an entity associated with the respective Asian culture can be placed. From these, we constructed our masked contexts by replacing the entity mentioned in the tweet with a \texttt{[MASK]} token. Similarly, to construct the culturally-neutral contexts of \texttt{Camellia-Neutral}, we identified tweets where entities from any culture would be appropriate as  \texttt{[MASK]}. 

\paragraph{Sentiment Annotation.} We annotated each context with one of three sentiment labels: \textit{positive}, \textit{negative}, or \textit{neutral}. This helps evaluate whether substituting the \texttt{[MASK]} token with the respective Asian or Western entities changes the sentiment predicted by LLMs (\S\ref{subsec:sentiment-association}).

\begin{figure*}[t]
    \centering
    \includegraphics[width=\linewidth]{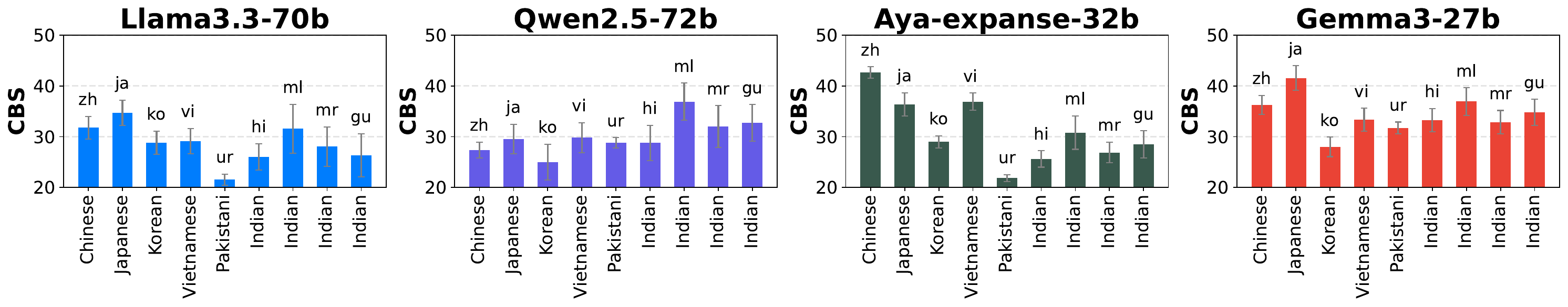}
    \caption{Average \textbf{C}ultural \textbf{B}ias \textbf{S}core (CBS) ($\downarrow$) across entity types achieved by LLMs on culturally-grounded contexts (\texttt{Camellia-Grounded}). LLMs can struggle to generate the appropriate Asian entities in each culture, assigning better likelihood to Western entities 30-40\% of the time.  See results per entity type in Appendix~\ref{appendix:results-cultural-adaptation}.}
    \label{fig:cbs-english}
\end{figure*}

\begin{figure*}[t]
    \centering
    \includegraphics[width=\linewidth]{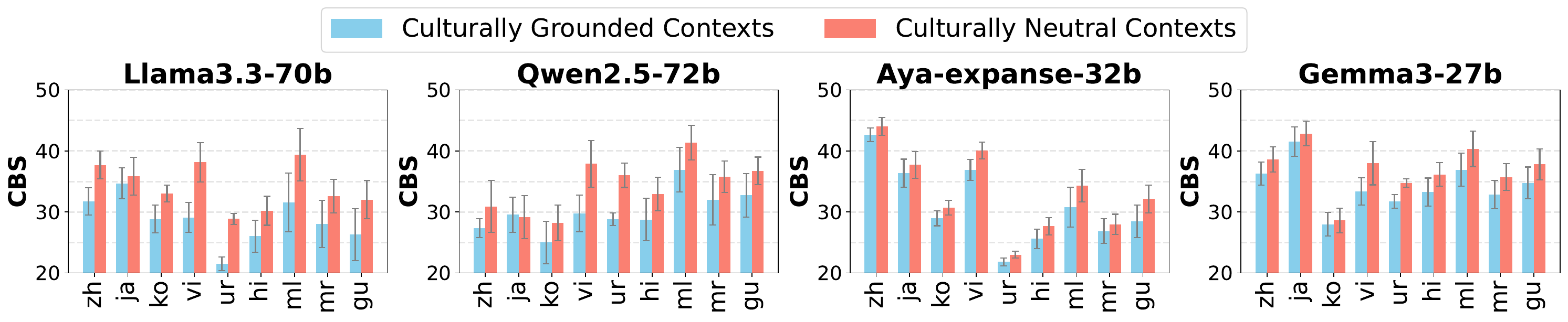}
    \caption{Average CBS across entity types on culturally-grounded contexts (\texttt{Camellia-Grounded}) vs culturally-neutral contexts (\texttt{Camellia-Neutral}) . LLMs show more preference towards Western entities in culturally-neutral contexts (higher CBS). CBS scores are lower in culturally-grounded contexts, yet remain close to the neutral case.}
    \label{fig:cbs-grounded-vs-neutral}
\end{figure*}

\paragraph{Contexts for Extractive QA.} In addition to the contexts used to evaluate cultural adaptation in LLMs, we constructed longer, paragraph-level contexts (Table~\ref{tab:contexts-stats-detailed}) in which entities are mentioned implicitly. These longer contexts enable a challenging evaluation setup for entity extraction, as they require understanding of the underlying context to identify the entity. We follow the same keyword search strategy to identify such contexts, and replace the mentioned entity with the \texttt{[MASK]} token. \texttt{Camellia-QA} provides $\sim$8-10 of such contexts for each entity type in each language.

\paragraph{Parallelizing Indian Contexts.} The contexts in \texttt{hi}, \texttt{ml}, \texttt{mr}, and \texttt{gu} were originally collected independently for each language. To enable comparisons across these Indian languages, we parallelized them by first translating the contexts into English and then into the other Indian languages.

\section{Are Cultural Biases Consistent Across Languages and LLMs?}
\label{sec:experiments}

We leverage the cultural entities and masked contexts in \texttt{Camellia} to investigate whether cultural biases are persistent across languages and LLMs.

\paragraph{Models.} We experiment with four open-weight LLMs with multilingual abilities: \textbf{Llama3.3-70b} \cite{grattafiori2024llama}, \textbf{Qwen2.5-72b} \cite{qwen2025qwen25technicalreport}, \textbf{Aya-expanse-32b} \cite{dang2024aya}, and \textbf{Gemma3-27b} \cite{team2025gemma}.

\paragraph{Tasks.} We test the LLMs under three setups: \textbf{cultural context adaptation} (assigning probabilities to the culturally relevant entities in cultural contexts; \S\ref{subsec:cbs}), \textbf{sentiment association} (sensitivity to changes of cultural entities within the same contexts when predicting sentiment; \S\ref{subsec:sentiment-association}), and \textbf{entity extractive QA} (ability to identify entities from different cultures within the same paragraph contexts; \S\ref{subsec:extractive-qa}). Additional experimental details and results are provided in Appendices~\ref{appendix:experimental-details} and ~\ref{appendix:additional-results}, respectively.

\subsection{Cultural Context Adaptation}
\label{subsec:cbs}

We first assess the ability of LLMs to adapt to different Asian cultural contexts by analyzing their assigned likelihood for the respective Asian vs Western entities as \texttt{[MASK]} token fillings.

\paragraph{Cultural Bias Score (CBS).} We use the CBS \cite{naous2024having} to measure the level of Western bias in an LLM$_{\theta}$. CBS is a likelihood-based measure that computes the percentage of an LLM's preference for Western entities over Asian ones within the same cultural context. Given an entity type $D$, two type-specific sets of respective Asian entities $ A = \{a_i\}_{i=1}^{N}$ and Western entities $B = \{b_j\}_{j=1}^{M}$, and a masked context $c_k$, we compute $\mathrm{CBS}_{D}(\text{LLM}_{\theta}, A, B, c_k)$ per language as:

\begin{equation}
    \frac{1}{N \times M} \sum_{i=1}^{N} \sum_{j=1}^M \mathbbm{1} [  P_{\mathtt{[MASK]} }(b_j|c_k)  > P_{\mathtt{[MASK]} }(a_i|c_k) ]
\end{equation}

\noindent where $P_{\texttt{[MASK]}}$ is the LLM's probability of an entity filling the \texttt{[MASK]} token. For entities tokenized into multiple tokens, we take the product of the conditional probabilities of each token. For a set of prompts $C = \{c_k\}_{k=1}^{K}$, the CBS per entity type for an LLM is computed by averaging over all $c_k \in C$. An LLM is considered more Western-biased as its CBS gets close to 100\%.

\begin{figure*}[t]
    \centering
    \includegraphics[width=0.9\linewidth]{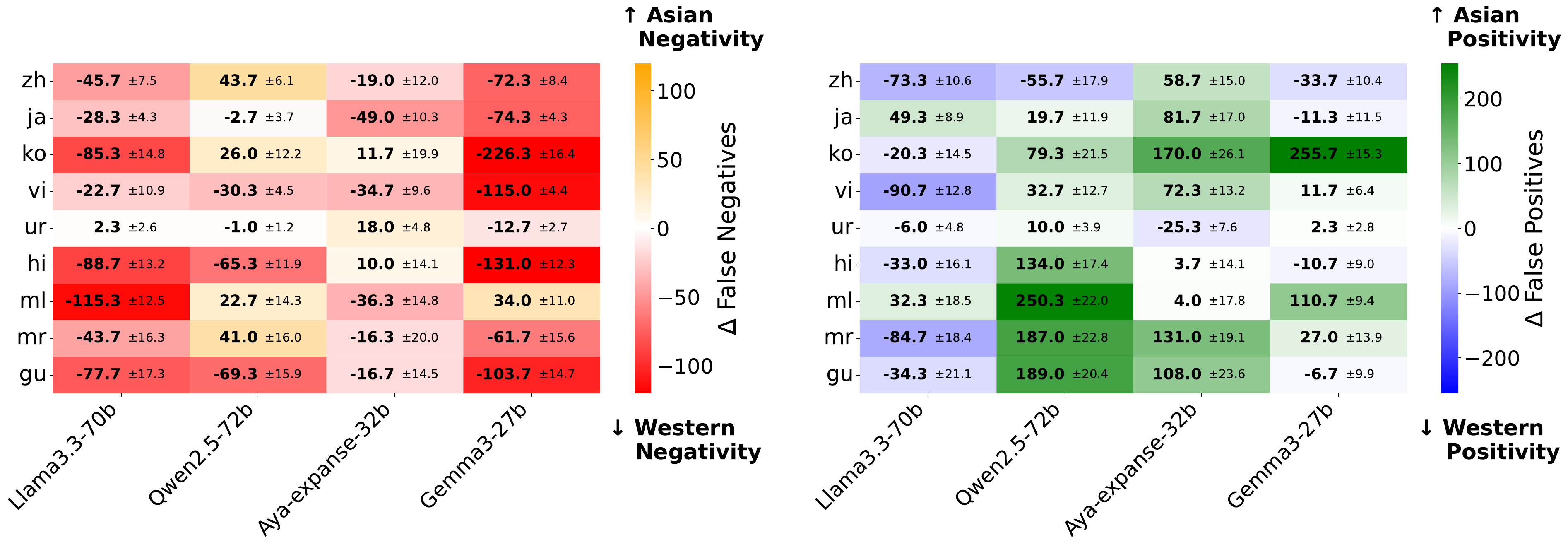}
    \caption{Differences in False Negative (FN) and False Positive (FP) sentiment predictions by LLMs on \texttt{Camellia} contexts filled with Asian vs Western entities. Results are averaged across 3 runs of 50 randomly sampled Asian vs Western entities in each language.}
    \label{fig:sentiment-results}
\end{figure*}

\paragraph{Results.}  Figures~\ref{fig:cbs-english} and ~\ref{fig:cbs-grounded-vs-neutral} show the average CBS across entity types when tested in each language. The scores are averaged across 3 runs where 50 Western and Native entities are sampled in each context across all entity types. We observe the following key insights:

\paragraph{LLMs can struggle to distinguish Asian vs. Western entities.}  Since the contexts we test on are grounded in each Asian culture (only entities associated with the specific Asian culture are appropriate for filling the \texttt{[MASK]}), models should always assign higher likelihood to the native Asian entities in those contexts, and the CBS is expected to be low (closer to the 0-5\% range). However, as Figure~\ref{fig:cbs-english} shows, in most cases the CBS is in the 30-40\% range. This highlights a large number of cases where LLMs struggle to differentiate between Asian and Western entities, assigning a higher likelihood to Western ones despite them being inappropriate to the context.

\paragraph{Are models sensitive to cultural grounding?} We further analyze if performance changes when testing on the contexts that are culturally neutral (i.e., any entity is an appropriate \texttt{[MASK]} filling in the context). The results in Figure~\ref{fig:cbs-grounded-vs-neutral} show that CBS scores are slightly higher when contexts are neutral. In the majority of cases, the scores remain very close to when contexts are culturally grounded. This suggests a lack of sensitivity to cultural contexts in LLMs, whereby their ability to select the appropriate entities at generation time is not greatly impacted by cultural grounding.

\paragraph{Adaptation ability varies by LLM family.} Noticeable differences can be seen in the performance of LLM families developed in different regions. Specifically, the Qwen2.5-72b model that is developed  by China-based Alibaba performs the best on Chinese, Japanese, and Korean, compared to the rest of the models. This corroborates the results of past work that shows a better ability of Qwen models at answering questions specific to Chinese culture \cite{guo2025care}. One likely reason for such a gap could be varying access to culturally relevant pre-training data. Because the pre-training datasets of these open-weight models are not publicly disclosed, directly analyzing differences in their training data remains challenging. However, we present an additional analysis in Appendix~\ref{appendix:results-cultural-adaptation}, where we compare model tokenizers, which offer a lens into their pre-training data characteristics \cite{hayase2024data}. Our results indicate that greater representation of a language’s script in a model’s tokenizer, which suggests stronger data coverage during frequency-based vocabulary construction, is associated with improved performance.

\begin{figure*}[t]
    \centering
    \includegraphics[width=\linewidth]{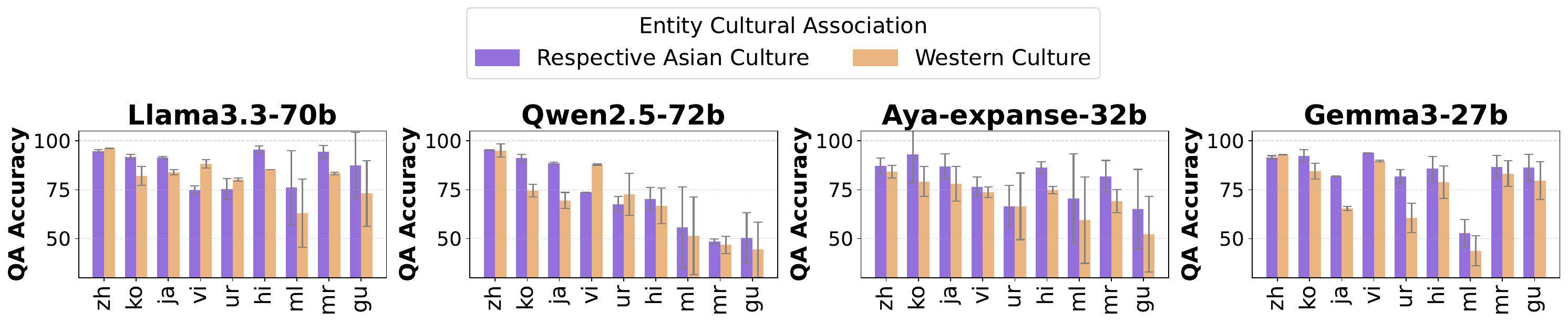}
    \caption{Extractive QA accuracy by LLMs on \texttt{Camellia-QA} contexts containing Asian vs Western entities when tested in each Asian language. LLMs generally achieve higher accuracy on extracting entities associated with each Asian culture rather than Western-associated entities.}
    \label{fig:results-qa}
\end{figure*}

\subsection{Sentiment Association}
\label{subsec:sentiment-association}

We now examine whether LLMs subtly associate entities from Asian or Western cultures with specific sentiment labels.

\paragraph{Setup.} We leverage the masked contexts in \texttt{Camellia-Grounded} and \texttt{Camellia-Neutral} that were manually annotated for sentiment to create a test set in each language. For each context, we replace the \texttt{[MASK]} token with 50 randomly sampled culture-specific Asian and Western entities. This results in two separate evaluation sets of $\sim$20k sentences per language: one with culture-specific Asian entities and the other with Western entities. Importantly, the contexts remain the same across both sets, allowing us to isolate the effect of entity cultural association on changes in the LLMs' predictions. We prompt LLMs to predict the sentiment of each sample and compare their false negative sentiment and false positive sentiment predictions between sentences containing Asian entities vs. Western entities. Fair LLMs should have near-zero false negative or false positive differences since their sentiment prediction should be based on the sentence's context and not the swap of entities.

\paragraph{Results.} Figure~\ref{fig:sentiment-results} shows the average differences in false negative and false positive predictions by LLMs for each language. We find that sentiment associations vary greatly across different LLMs. Llama and Gemma generally show a stronger tendency to associate Western entities with negative sentiment, whereas Qwen often associates Asian entities with positive sentiment, particularly in Indian languages. These results highlight how current LLMs can be sensitive to cultural associations of entities leading to biased misclassifications.

\subsection{Entity Extractive QA}
\label{subsec:extractive-qa}

Finally, we analyze the ability of LLMs to extract entities from paragraph-long contexts. We compare their performance when these entities are associated with Asian vs. Western cultures.

\paragraph{Setup.} Using the contexts from \texttt{Camellia-QA}, we construct Asian and Western test sets in each language. For each context, we replace the \texttt{[MASK]} with 50 randomly sampled entities, in a similar manner to our earlier experiment for sentiment association (\S\ref{subsec:sentiment-association}). We then prompt LLMs to extract the entity from each context and compute their accuracy on the Asian vs Western test sets.

\begin{table}[t]
\centering
\resizebox{1.0\linewidth}{!}{%
\begin{tabular}{@{}lcc|cc|cc|cc@{}}
\cmidrule(l){2-9}
 & \multicolumn{2}{c|}{\textbf{Llama3.3-70b}} & \multicolumn{2}{c|}{\textbf{Qwen2.5-72b}} & \multicolumn{2}{c|}{\textbf{Aya-expanse-32b}} & \multicolumn{2}{c}{\textbf{Gemma3-27b}} \\ \cmidrule(l){2-9} 
Culture & \textbf{Asian} & \textbf{English} & \textbf{Asian} & \textbf{English} & \textbf{Asian} & \textbf{English} & \textbf{Asian} & \textbf{English} \\ \midrule
Chinese & \cellcolor[HTML]{DCF3F4}-1.32 & \cellcolor[HTML]{FFF3EA}0.30 & \cellcolor[HTML]{DCF3F4}0.43 & \cellcolor[HTML]{FFF3EA}-2.84 & \cellcolor[HTML]{DCF3F4}2.84 & \cellcolor[HTML]{FFF3EA}-5.83 & \cellcolor[HTML]{DCF3F4}-1.36 & \cellcolor[HTML]{FFF3EA}-5.63 \\
Japanese & \cellcolor[HTML]{98DBE0}7.55 & \cellcolor[HTML]{FFF3EA}2.72 & \cellcolor[HTML]{46BDC6}18.87 & \cellcolor[HTML]{FFF3EA}4.53 & \cellcolor[HTML]{80D2D8}8.84 & \cellcolor[HTML]{FFF3EA}-0.73 & \cellcolor[HTML]{46BDC6}16.40 & \cellcolor[HTML]{FFF3EA}-3.22 \\
Korean & \cellcolor[HTML]{98DBE0}9.69 & \cellcolor[HTML]{FFF3EA}0.66 & \cellcolor[HTML]{46BDC6}16.47 & \cellcolor[HTML]{FFF3EA}-2.49 & \cellcolor[HTML]{46BDC6}13.94 & \cellcolor[HTML]{FFF3EA}1.43 & \cellcolor[HTML]{80D2D8}7.94 & \cellcolor[HTML]{FFF3EA}2.54 \\
Vietnamese & \cellcolor[HTML]{46BDC6}-13.53 & \cellcolor[HTML]{FFF3EA}1.95 & \cellcolor[HTML]{46BDC6}-14.33 & \cellcolor[HTML]{FFF3EA}-3.61 & \cellcolor[HTML]{DCF3F4}2.83 & \cellcolor[HTML]{FFF3EA}-1.88 & \cellcolor[HTML]{DCF3F4}4.15 & \cellcolor[HTML]{FFF3EA}1.65 \\
Pakistani & \cellcolor[HTML]{DCF3F4}-4.71 & \cellcolor[HTML]{FFC092}10.54 & \cellcolor[HTML]{DCF3F4}-4.99 & \cellcolor[HTML]{FFC092}12.16 & \cellcolor[HTML]{DCF3F4}0.12 & \cellcolor[HTML]{FFF3EA}4.54 & \cellcolor[HTML]{46BDC6}21.11 & \cellcolor[HTML]{FFF3EA}4.54 \\
Indian \texttt{(hi)} & \cellcolor[HTML]{46BDC6}10.05 & \cellcolor[HTML]{FFDDC4}6.71 & \cellcolor[HTML]{DCF3F4}3.63 & \cellcolor[HTML]{FFC092}10.67 & \cellcolor[HTML]{46BDC6}11.54 & \cellcolor[HTML]{FFF3EA}1.07 & \cellcolor[HTML]{80D2D8}6.81 & \cellcolor[HTML]{FFF3EA}3.25 \\
Indian \texttt{(ml)} & \cellcolor[HTML]{46BDC6}13.15 & --- & \cellcolor[HTML]{DCF3F4}4.22 & --- & \cellcolor[HTML]{46BDC6}10.93 & --- & \cellcolor[HTML]{80D2D8}9.01 & --- \\
Indian \texttt{(mr)} & \cellcolor[HTML]{46BDC6}11.07 & --- & \cellcolor[HTML]{DCF3F4}1.68 & --- & \cellcolor[HTML]{46BDC6}12.64 & --- & \cellcolor[HTML]{DCF3F4}3.50 & --- \\
Indian \texttt{(gu)} & \cellcolor[HTML]{46BDC6}14.44 & --- & \cellcolor[HTML]{80D2D8}6.02 & --- & \cellcolor[HTML]{46BDC6}12.89 & --- & \cellcolor[HTML]{80D2D8}6.54 & --- \\ \bottomrule
\end{tabular}
}
\caption{$\Delta$Accuracy on extractive QA between Western and Asian entities when testing models on parallel data in the respective Asian language of each culture vs. in English. Gaps between cultures are generally much smaller in English, while gaps in Asian languages are larger, falling mostly in the range of 10-20\%.}
\label{tab:qa-asian-english-comparison}
\end{table}

\paragraph{Results.} Figure~\ref{fig:results-qa} shows the accuracies achieved by LLMs. We find many cases where LLMs generally achieve higher accuracy in extracting the native Asian entities rather than Western ones. A few languages show the opposite behavior, specifically in Vietnamese and Urdu, where Llama and Qwen achieve higher accuracy on Western entities.

To compare whether similar gaps also appear in English, we evaluate all models on parallel English data for each culture. The results are shown in Table~\ref{tab:qa-asian-english-comparison}. In English, cultural gaps are generally small, mostly between 1\% and 5\%, with no consistent advantage for either culture. In contrast, gaps in Asian languages are often larger, reaching 12\%-20\%, except in Chinese where the differences remain minimal. These results suggest that LLMs still struggle to capture implicit cultural context in many non-English languages, leading to substantially larger performance disparities across cultures.

\section{Conclusion}
We introduced \texttt{Camellia}, a benchmark for evaluating entity-centric cultural biases in 9 Asian languages across 6 Asian cultures. Through various evaluations, we showed that current multilingual LLMs exhibit various types of cultural biases in these non-Western languages which can manifest in poor cultural adaptation, biased sentiment associations, and accuracy gaps in entity extraction tasks. We hope that \texttt{Camellia} will serve as a valuable resource to support future research aimed at developing more culturally fair multilingual LLMs.

\section*{Limitations}
In \texttt{Camellia}, we defined the broad Western culture as countries that are exclusively in North America and Europe. However, there are several countries in other geographical regions where Western culture dominates such as Australia, New Zealand, and South American countries, that were excluded from our definition. Our focus was to explore cultural biases in LLMs when contrasting entities associated each Asian culture we study against those associated with the broad Western culture. We thus followed the view of North America and Europe as representing Western culture, and for which data could be more easily collected in the Asian languages we study. Future work can expand on our set of Western entities to include more representation from these regions to enable more fine-grained comparisons.

\section*{Ethics Statement}

While collecting data from social media posts to construct the masked contexts in \texttt{Camellia}, we discarded any tweets that included offensive or toxic language, hate speech, stereotypes, or included any personally identifiable information. Data collection was done through a manual process by searching on the X, Weibo, and Xiaohongshu  platforms, without the use of automated scraping. We do not share the raw social media posts but modified, slightly re-written versions where cultural entities are replaced by a \texttt{[MASK]}, which can be used for research purposes. \texttt{Camellia} is constructed for the purpose of testing cultural biases in LLMs and enabling future research on the development of LLMs that work efficiently and fairly for all entities regardless of the cultural associations they carry.

\section*{Reproducibility Statement}

The \texttt{Camellia} benchmark will be made publicly available to the community, which includes the collected entities with their annotations for cultural association and the masked contexts for all languages. We provide in Appendix~\ref{appendix:camellia-details} the annotation guideline we used to annotate entities, and additional experimental details in Appendix~\ref{appendix:experimental-details}, such as the prompts and decoding configurations that can be used to replicate our experiments for all languages.

\section*{Acknowledgments}

The authors would like to thank Sara Takagi, Huong-Tra Le-Nguyen, Kiran Khan, and Ehab Ashraf for their help with data annotation, and Xiaofeng Wu for performing post-annotation quality-checks. Tarek Naous, Geyang Guo, Alan Ritter, and Wei Xu are partially supported by the U.S. National Science Foundation CAREER Awards IIS-2144493 and IIS-2052498. Any opinions,
findings, and conclusions or recommendations expressed in this material are those of the authors and
do not necessarily reflect the views of the National Science Foundation. 
Keisuke Sakaguchi is supported by the Nakajima Foundation and JSPS KAKENHI Grant Number JP25K03175. Jaehyeok Lee, Kyungdon Lee, and JinYeong Bak are partially supported by Institute of Information \& communications Technology Planning \& Evaluation (IITP) grant funded by the Korea government (MSIT) (No. RS-2024-00509258 and No. RS-2024-00469482, Global AI Frontier Lab). Tanmoy Chakraborty acknowledges the financial support of Anusandhan National Research Foundation (ANRF/ARG/2025/001505/QSS). Sarah Masud was supported by the Pioneer Centre for AI, DNRF grant number P1.

\section*{Dataset Contribution Statement}
{\footnotesize
\scalebox{1}{
\begin{tabular}{r |@{\foo} l}
\textbf{Chinese Data} & [Mengyu Ye, Geyang Guo]\\
\textbf{Japanese Data} & [Mengyu Ye]\\
\textbf{Korean Data} & [Kyungdon Lee, Jaehyeok Lee]\\
\textbf{Vietnamese Data} & [Trung Thanh Tran]\\
\textbf{Urdu Data} & [Zohaib Khan]\\
\textbf{Hindi Data} & [Sarah Masud, Sahajpreet Singh]\\
\textbf{Malayalam Data} & [Anagha Savit]\\
\textbf{Marathi Data} & [Tanish Patwa]\\
\textbf{Gujarati Data} & [Neel Kothari]\\
\end{tabular}
}
}

\bibliography{references}

\clearpage
\newpage

\appendix

\section{Camellia: Additional Details}
\label{appendix:camellia-details}

\subsection{Entity Statistics.} Table~\ref{tab:entities-stats-detailed} shows the number of entities for each language and entity type that we collect and annotate in \texttt{Camellia.}

\subsection{Wikidata Classes.} Table~\ref{tab:wikidata-classes} lists the Wikidata classes we used to extract cultural entities. For each language, we identify the relevant country (e.g., India for \texttt{hi, ml, gu} and \texttt{mr}, Pakistan for \texttt{ur}, Vietnam for \texttt{vi}, etc.) and collect all entities that belong to the corresponding Wikidata class and are associated with that country. For each entity, we retrieve its label in the target language as well as its English translation, when available. To collect Western entities, we extract entities for all countries in North America and Western Europe.

\subsection{Language-Specific Challenges}

We discuss some of the entity-specific challenges we encountered while constructing \texttt{Camellia}. These challenges stem from diverse linguistic and cultural factors that shaped several dataset design choices. Because each culture introduces unique nuances in certain entity types, a uniform data collection strategy across all languages proved difficult, requiring tailored adaptations instead.

\paragraph{Entity naming conventions can be subject to temporal change.} In Korea, China, and Japan, modern names differ significantly from older ones \cite{barevsova2023tradition}. For instance, many Korean feminine names in the mid-20th century included elements like `\textit{suk}' (\begin{CJK}{UTF8}{mj}숙\end{CJK}) or `\textit{mi}' (\begin{CJK}{UTF8}{mj}미\end{CJK}), which symbolize purity and beauty, respectively. In contrast, contemporary names like `\textit{Seo-yun}' (\begin{CJK}{UTF8}{mj}서윤\end{CJK}) or `\textit{Ji-woo}' (\begin{CJK}{UTF8}{mj}지우\end{CJK}) reflect trend-driven preferences. Chinese names have similarly shifted over the last century, becoming shorter and more unique due to political and social factors \citep{ogihara2023historical}. Such temporal changes make it challenging to collect entities that are representative today. For example, the Korean, Chinese, and Japanese first names listed on Wikidata are largely outdated, with little to no contemporary usage. To reflect modern naming conventions, we used recent governmental statistical reports in Korea\footnote{\url{https://efamily.scourt.go.kr}} and China\footnote{ \href{https://app.mps.gov.cn/gdnps/pc/content.jsp?id=8349293}{2021 National Name Report}}. For Japanese, due to a lack of similar reports, we used a popular name generator\footnote{\url{https://namegen.jp}} to generate Japanese first names, which were then verified to be valid by the annotators.

\paragraph{Entity types can persist in everyday use in some cultures but not in others.} The Arabic CAMeL benchmark \cite{naous2024having} initially included a clothing entity type contrasting traditional Arab clothing with Western attire. However, extending this to other non-Western cultures proved challenging. For instance, in Pakistani culture, traditional garments such as the ``\textit{shalwar kameez}" remain a common part of everyday attire \cite{ranavaade2017study}. In contrast, for other Asian societies, including China and Japan, traditional clothing like the ``\textit{hanfu}" is now generally reserved for special occasions. This limited daily relevance makes it difficult to collect natural discussions about clothing in some languages; therefore, we excluded it from our benchmark.

\paragraph{The same entity type may need to be tailored to local cultural popularity.} The same entity type can carry different meanings depending on the culture, reflecting what people care about and discuss. This is illustrated by the sports clubs category in \texttt{Camellia}. We focused on sports that have a strong imprint in each culture. In Pakistan and India, for example, cricket holds significant importance and even influences political discourse between the two countries; accordingly, we collected cricket clubs as the sports club entities for these cultures. In contrast, for cultures in East and Southeast Asia, we focused on football as one of the most widely followed sports \cite{connell2018globalisation}. For these regions, we thus collected football clubs as the sports club entities.

\begin{table}[h]
\centering
\begin{adjustbox}{width=\linewidth}
\begin{tabular}{@{}lccccccc@{}}
\cmidrule(l){2-8}
 & \multicolumn{7}{c}{\textbf{\#Cultural Entities}} \\ \midrule
\textbf{Entity Type} & \texttt{zh} & \texttt{ja} & \texttt{ko} & \texttt{vi} & \texttt{ur} & \texttt{hi/ml/mr/gu} & \multicolumn{1}{l}{western} \\ \midrule
Authors & 165 & 260 & 602 & 24 & 44 & 207 & 370 \\
Beverage & 189 & 115 & 107 & 77 & 11 & 34 & 497 \\
Food & 415 & 635 & 416 & 374 & 75 & 605 & 436 \\
Locations & 1,000 & 817 & 1,260 & 90 & 196 & 181 & 382 \\
Names (M) & 906 & 503 & 899 & 251 & 334 & 651 & 588 \\
Names (F) & 1,123 & 523 & 886 & 151 & 163 & 563 & 587 \\
Sports & 116 & 354 & 266 & 51 & 17 & 165 & 849 \\ \midrule
\multicolumn{1}{r}{\textbf{Total}} & 3,914 & 3,207 & 4,436 & 1,018 & 840 & 2,406 & 3,709 \\ \bottomrule
\end{tabular}
\end{adjustbox}
\caption{Number of entities for each language and entity type in \texttt{Camellia}. Western entities are parallel across all languages. Each entity is also available as an English translation.}
\label{tab:entities-stats-detailed}
\end{table}

\begin{table}[h]
\centering
\begin{adjustbox}{width=\linewidth}
\begin{tabular}{@{}lll@{}}
\toprule
\textbf{Entity Type} & \textbf{Wikidata Class} & \textbf{Class QID} \\ \midrule
\multirow{2}{*}{Authors} & writer & Q36180 \\
 & novelist & Q6625963 \\ \midrule
Beverage & drink & Q40050 \\ \midrule
\multirow{2}{*}{Food} & food & Q2095 \\
 & dish & Q746549 \\ \midrule
Location & city & Q515 \\ \midrule
Names (F) & female given name & Q11879590 \\ \midrule
Names (M) & male given name & Q12308941 \\ \midrule
\multirow{2}{*}{Sports Clubs} & association football club & Q476028 \\
 & cricket team & Q17376093 \\ \bottomrule
\end{tabular}
\end{adjustbox}
\caption{Wikidata classes used to extracting entities for each entity type in all languages.}
\label{tab:wikidata-classes}
\end{table}

\begin{figure}[h]
    \centering
    \includegraphics[width=\linewidth]{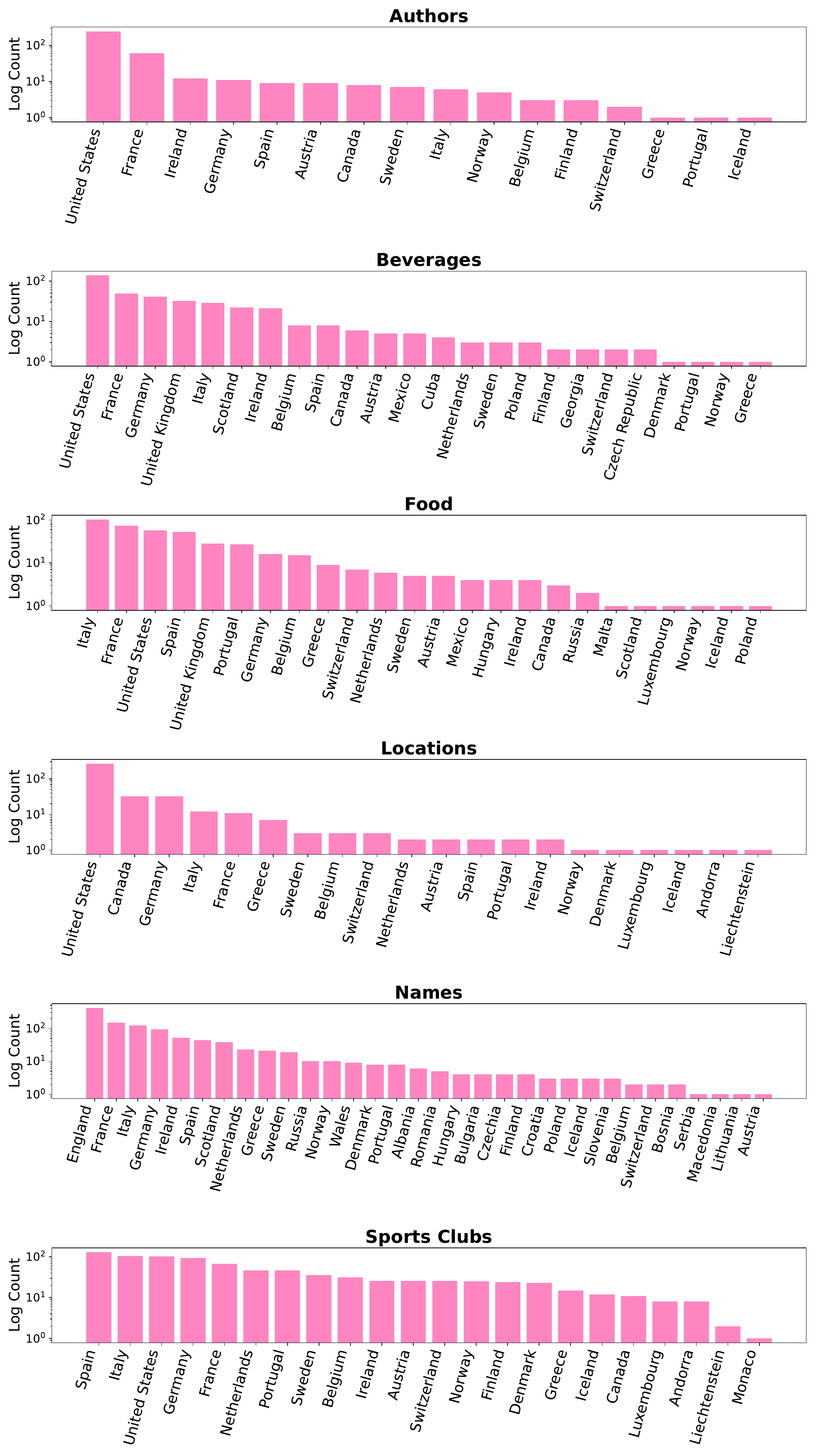}
    \caption{Country-wise distribution of Western entities in \texttt{Camellia} for different entity types.}
    \label{fig:western-entities-dist}
\end{figure}

\paragraph{Country distribution of Western entities.} Figure~\ref{fig:western-entities-dist} reports the country-wise distribution of Western entities in \texttt{Camellia}. The countries of origin for authors, beverage, food, locations, and sport clubs, and entities were obtained from Wikidata which provides country of origin label for most entities, with the exception of some food and beverage entities that we manually annotated for origin. For Western first names, we prompted GPT-4o to classify the origin of each name to obtain the distribution in that entity type. We then manually verified that all labels were accurate. Examples include: Panagiotis as Greek, Javienne as French, Marilo as Italian, Erling as Norwegian, etc. These country labels are only for visualization purposes and are not used in our experiments.

\paragraph{Typological Diversity.} The languages in \texttt{Camellia} represent a broad span of typological diversity in terms of genealogical families, writing systems, and morphological profiles. We summarize those in Table~\ref{tab:typo-diversity} and report some details of each language below: 

\begin{itemize}
    \item \textbf{Chinese:} Chinese is a Sino-Tibetan language with a highly isolating morphology. It uses a logographic writing system (Han characters) that primarily encodes morphemes but also incorporates phonetic components, making it distinct from alphabetic scripts in terms of structure. The language is also tonal, adding further phonological complexity.
    
    \item \textbf{Japanese}: Japanese belongs to the Japonic family and exhibits agglutinative morphology (i.e, grammatical markers attach transparently to stems). Its writing system is tri-scriptal, combining Kanji (logographic) with Hiragana and Katakana (syllabaries). 

    \item \textbf{Korean:} Korean is a Koreanic language. It uses Hangul, a featural alphabet whose letters combine into block-like syllabic units, creating a script that is alphabetic in design but syllabic in appearance. Korean is also agglutinative, with rich postpositional case particles and verbal morphology.

    \item \textbf{Vietnamese:} Vietnamese is an Austroasiatic language that is heavily shaped by historical Chinese contact. It has an analytic/isolating morphology with little inflection, and is tonal, distinguishing meaning through pitch contours. Its modern writing system is Latin-based but employs extensive diacritics for tones and vowel quality.

    \item  \textbf{Urdu:} Urdu is an Indo-Aryan language with fusional morphology, expressing multiple grammatical categories through single affixes. It is written in Perso-Arabic script, a right-to-left script with complex ligatures and highly variable glyph shapes. 

    \item \textbf{Hindi:} Hindi is an Indo-Aryan language that shares a lot of its grammatical structure with Urdu but differs in script. It has a fusional morphology, with rich agreement and case marking. Hindi uses the Devanagari  alphasyllabary.

    \item \textbf{Malayalam:} Malayalam is a Dravidian language characterized by agglutinative morphology and long, morphologically complex word forms. Its Malayalam alphasyllabary has a large inventory of characters and ligatures. 

    \item \textbf{Marathi:} Marathi is an Indo-Aryan language with fusional morphology and extensive nominal and verbal inflection. It is written in Devanagari but includes additional letters not found in Hindi, leading to differences in sound and usage.

    \item \textbf{Gujarati:} Gujarati is an Indo-Aryan language written in its own Gujarati alphasyllabary, which is historically related to but visually distinct from Devanagari. It exhibits fusional morphology with case marking, gender agreement, and verb inflection.  
    
\end{itemize}
It is interesting to note that among the languages we study, four are gendered: Urdu, Hindi, Marathi, and Gujarati.

\begin{table}[]
\centering
\begin{adjustbox}{width=\linewidth}
\begin{tabular}{@{}llll@{}}
\toprule
\textbf{Language} & \textbf{Family} & \textbf{Morphology} & \textbf{Script} \\ \midrule
\texttt{zh} & Sino-Tibetan & Isolating & Logographic \\
\texttt{ja} & Japonic & Agglutinative & Logographic \& Syllabic \\
\texttt{ko} & Koreanic & Agglutinative & Alphabetic (Hangul) \\
\texttt{vi} & Austroasiatic & Analytic/Isolating & Latin \\
\texttt{ur} & Indo-Aryan & Fusional & Perso-Arabic Nastaliq \\
\texttt{hi} & Indo-Aryan & Fusional & Alphasyllabary (Devanagari) \\
\texttt{ml} & Dravidian & Agglutinative & Alphasyllabary (Malayalam) \\
\texttt{mr} & Indo-Aryan & Fusional & Alphasyllabary (Devanagari) \\
\texttt{gu} & Indo-Aryan & Fusional & Alphasyllabary (Gujarati) \\ \bottomrule
\end{tabular}
\end{adjustbox}
\caption{Typological Diversity of the languages in Camellia.}
\label{tab:typo-diversity}
\end{table}

\subsection{Annotation Guideline} 
\begin{figure*}[h]
    \centering
    \fbox{
        \includegraphics[width=0.95\linewidth]{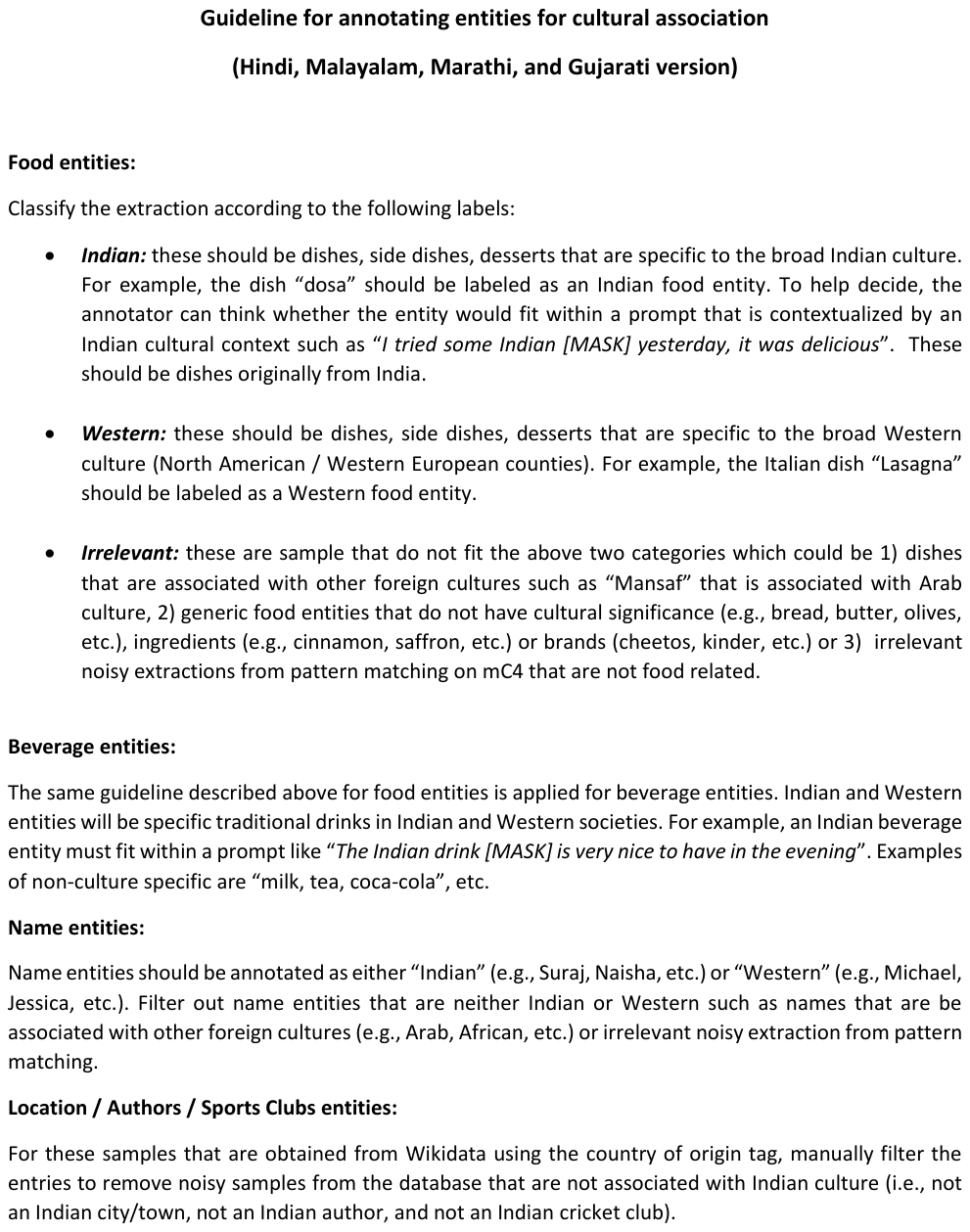}
    }
    \caption{Indian-focused version of our annotation guideline for annotating cultural entities.}
    \label{fig:annotation-guideline}
\end{figure*}

Figure~\ref{fig:annotation-guideline} shows our guideline for annotating cultural entities across all entity types, focusing on Indian culture for Hindi, Malayalam, Marathi, and Gujarati. We adapted the guideline for the other cultures/languages by switching examples as necessary.

\subsection{Examples of Culturally-Grounded Contexts.} 
\begin{figure*}[h]
    \centering
    \fbox{
        \includegraphics[width=\linewidth]{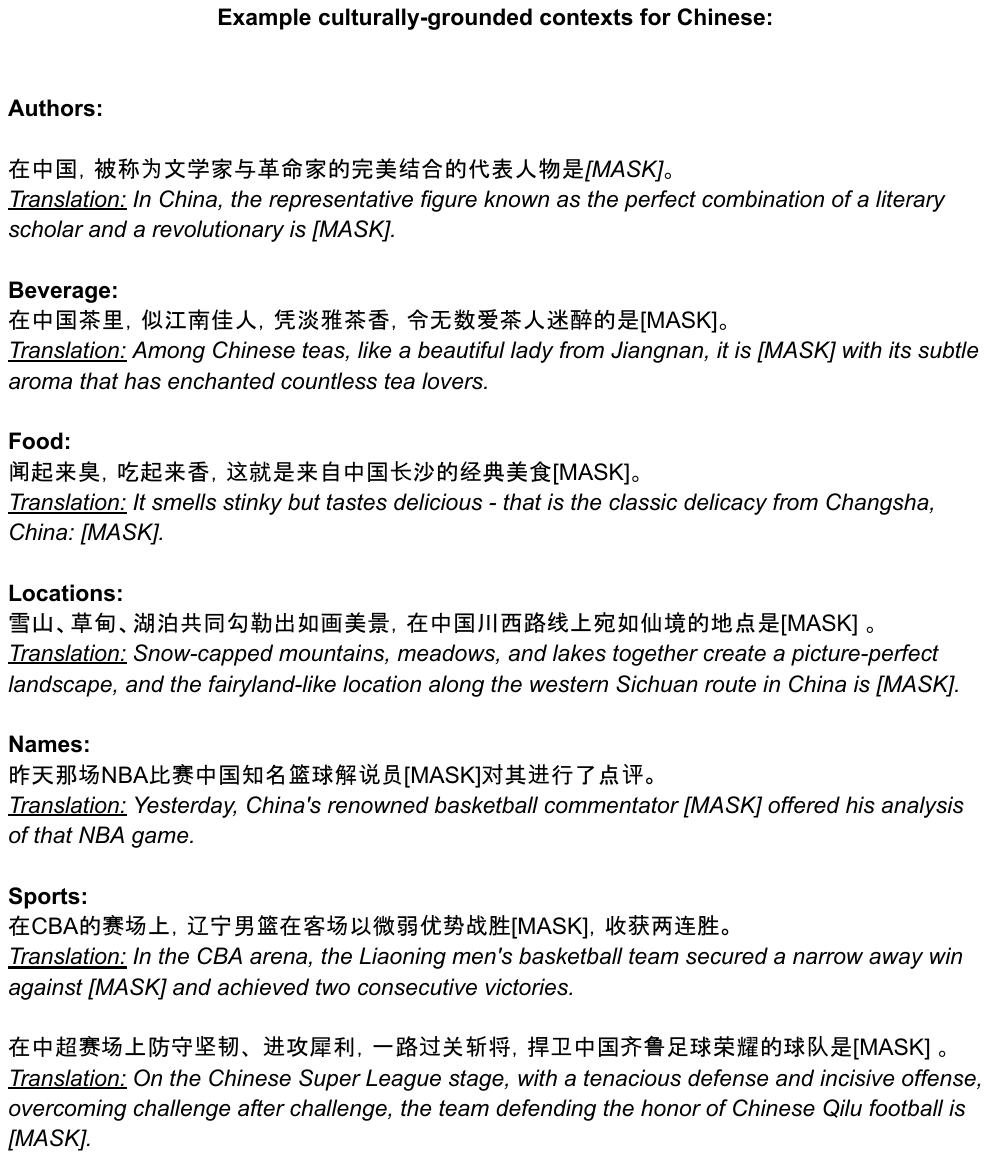}
    }
    \caption{Examples of culturally-grounded masked contexts for Chinese culture from \texttt{Camellia-Grounded}.}
    \label{fig:chinese-grounded-examples}
\end{figure*}

Figure~\ref{fig:chinese-grounded-examples} shows examples of culturally-grounded masked contexts for Chinese culture from \texttt{Camellia-Grounded}. In these examples, only entities associated with Chinese culture would be appropriate to fit the \texttt{[MASK]}.

\subsection{Examples of Culturally-Neutral Contexts.} 
\begin{figure*}[h]
    \centering
    \fbox{
        \includegraphics[width=\linewidth]{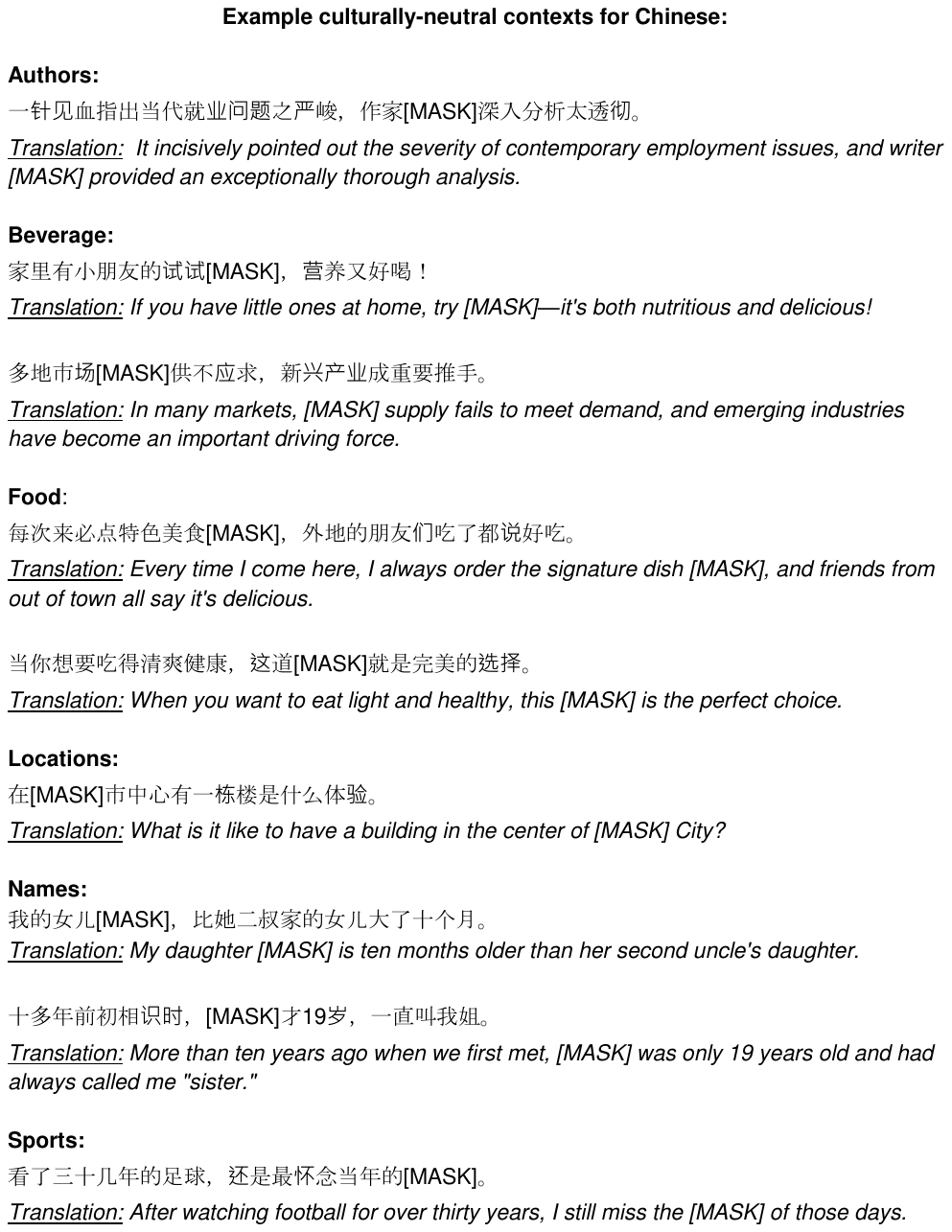}
    }
    \caption{Examples of culturally-neutral masked contexts for Chinese culture from \texttt{Camellia-Neutral}.}
    \label{fig:chinese-neutral-examples}
\end{figure*}

Figure~\ref{fig:chinese-neutral-examples} shows examples of culturally-neutral masked contexts for Chinese culture from \texttt{Camellia-Neutral}. In these examples, entities associated with any culture would be appropriate to fit the \texttt{[MASK]}.

\clearpage
\newpage

\section{Additional Experimental Details}
\label{appendix:experimental-details}

\paragraph{Prompts for extractive QA and sentiment classification.} We used the same prompt used by \citet{naous2024having} for our sentiment association experiment, where models are given a key and asked to classify the sentiment of the given sentence (see Table~\ref{tab:sentiment-prompt}). We also used the prompt by \citet{naous-xu-2025-origin} for the extractive QA experiment, where models are given the context and entity type we seek to extract asked to identify the entity mentioned in the text (see Table~\ref{tab:qa-prompt}).

\paragraph{Inference Details and Parameters.} We ran our experiments using 8 NVIDIA A40 GPUs. We used the vLLM library\footnote{\url{https://docs.vllm.ai}} \cite{kwon2023efficient} for fast inference on the extractive QA and sentiment association tasks in each language. Greedy decoding was selected by setting the following parameters \{\texttt{temperature=0}, \texttt{top\_p=1}, \texttt{top\_k=1}\}. We limited the number of generated tokens by the models by setting \{\texttt{max\_tokens=30}\}. We also set the context length to \{\texttt{max\_model\_len=4096}\}, which fit all of the contexts in our benchmark.

\begin{table}[]
\centering
\begin{adjustbox}{width=\linewidth}
\begin{tabular}{@{}l@{}}
\toprule
\begin{tabular}[c]{@{}l@{}}

\texttt{Classify the sentiment in this \{LANGUAGE\} sentence based on} \\
\texttt{the following key:} \\
\\
\texttt{0 = neutral} \\ 
\texttt{1 = positive} \\ 
\texttt{2 = negative} \\ 
\\

\texttt{Sentence: ``\{SENTENCE\}''}\\
\texttt{Given the above key, the sentiment of this sentence is (0-2):} 
\end{tabular} \\
\bottomrule
\end{tabular}
\end{adjustbox}
\caption{Prompt used to classify a sentence's sentiment in our sentiment association experiment.}
\label{tab:sentiment-prompt}
\end{table}

\begin{table}[]
\centering
\begin{adjustbox}{width=\linewidth}
\begin{tabular}{@{}l@{}}
\toprule
\begin{tabular}[c]{@{}l@{}}

\texttt{Extract the \{ENTITY\_TYPE\} entity mentioned in the} 
\\\texttt{following \{LANGUAGE\} text.} \\

\\
\texttt{Text: ``\{QA\_CONTEXT\}''}\\
\\
\texttt{Reply only with the mentioned \{ENTITY\_TYPE\}.} \\
\texttt{If nothing is found, reply ``None''.} 

\end{tabular} \\
\bottomrule
\end{tabular}
\end{adjustbox}
\caption{Prompt used to extract entities from contexts in our extractive QA experiment.}
\label{tab:qa-prompt}
\end{table}

\clearpage
\newpage

\begin{figure*}[t!]
    \centering
    \includegraphics[width=\linewidth]{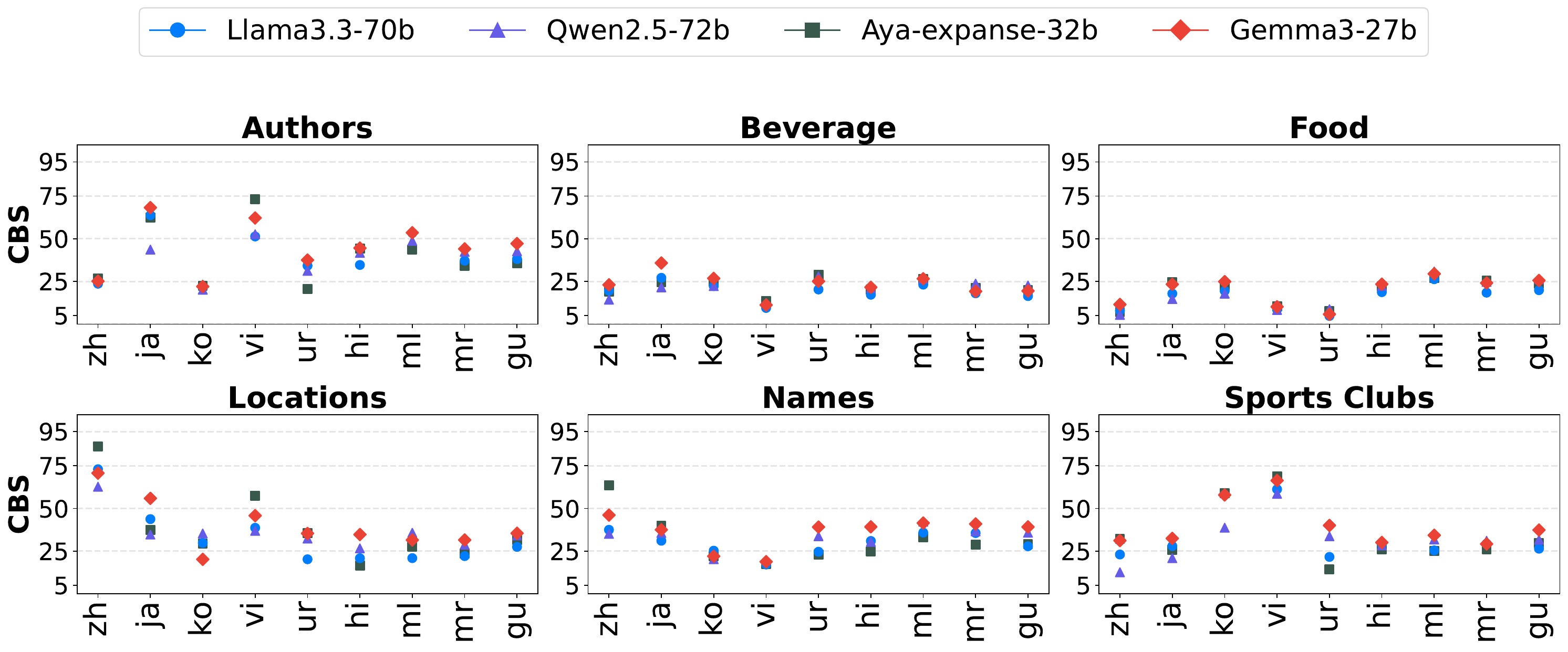}
    \caption{\textbf{C}ultural \textbf{B}ias \textbf{S}core (CBS) ($\downarrow$) (\S\ref{subsec:cbs}) per entity type achieved by LLMs on culturally-grounded contexts (\texttt{Camellia-Grounded}) for each Asian language. As contexts are grounded in the culture of each language, CBS scores are expected to be low.}
    \label{fig:cbs-camellia-co}
\end{figure*}

\begin{figure*}[t]
    \centering
    \includegraphics[width=\linewidth]{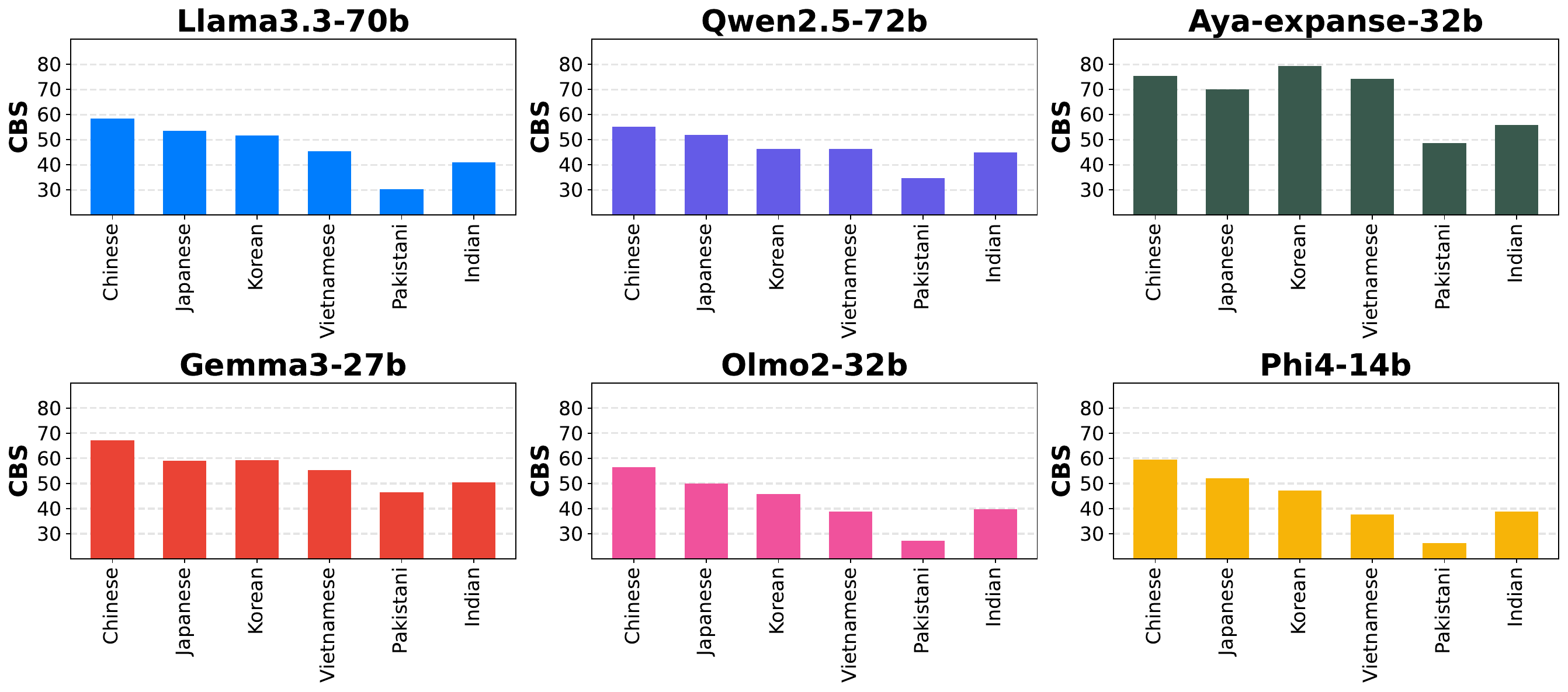}
    \caption{Average \textbf{C}ultural \textbf{B}ias \textbf{S}core (CBS) ($\downarrow$) across entity types achieved by LLMs on culturally-grounded contexts (\texttt{Camellia-Grounded}) when tested in English for each culture.}
    \label{fig:cbs-camellia-co-english}
\end{figure*}

\section{Additional Results}
\label{appendix:additional-results}

\subsection{Cultural Adaptation}
\label{appendix:results-cultural-adaptation}

\paragraph{CBS scores per Entity Type.} Figure~\ref{fig:cbs-camellia-co} shows the CBS per entity-type achieved by LLMs when tested on the culturally-grounded contexts. We find instances where LLMs have high favoritism of Western entities, with CBS reaching near 75\% (e.g., authors in \texttt{vi} and \texttt{ja}). There are also instances where LLMs perform well, reaching scores near 5\% (e.g., food entities in \texttt{zh}, and \texttt{ur}).

\paragraph{CBS scores when testing in English.} Figure~\ref{fig:cbs-camellia-co-english} shows the average CBS achieved by each model on the culturally-grounded contexts in \texttt{Camellia} when tested on the English translations. Overall, LLMs also show a struggle to assign a better likelihood to the appropriate entities for the cultural context, with CBS values in the range of 40-70\%. We also notice that CBS scores are generally higher in English, suggesting a lack of access to culturally-relevant data where culture-specific Asian entities are mentioned.

\begin{figure*}[t]
    \centering
    \includegraphics[width=\linewidth]{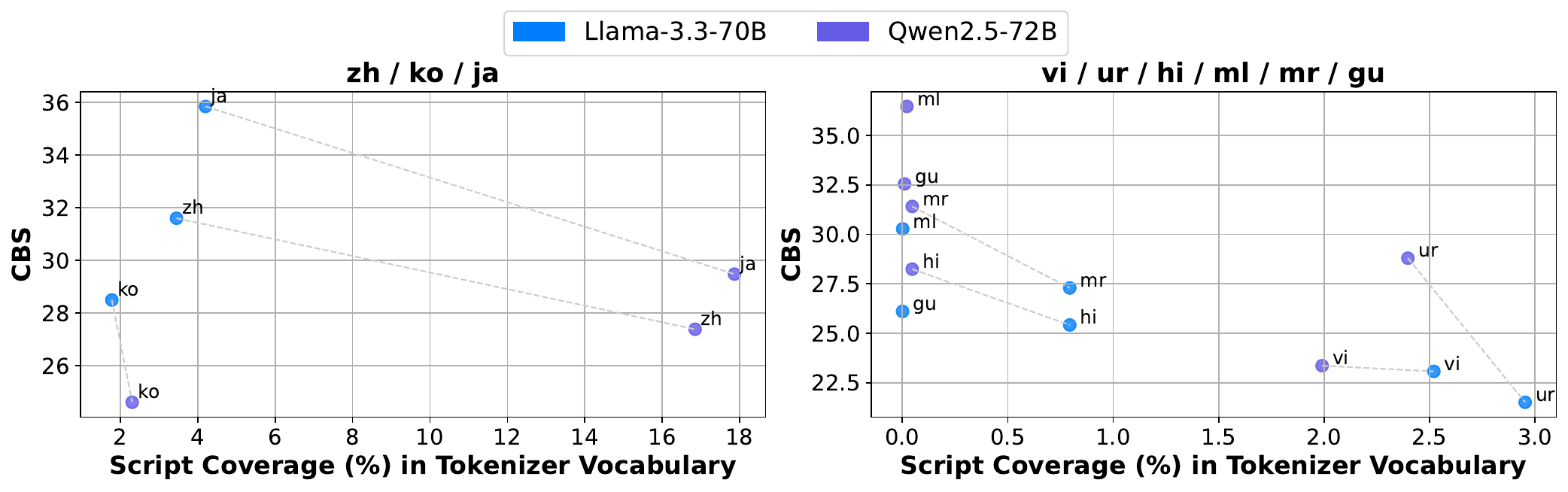}
    \caption{CBS vs script coverage \% in tokenizer vocabulary for Llama3.3-70b and Qwen2.5-72b. Higher script coverage in a tokenizer tends to yield better performance (i.e., lower CBS). Dashed gray lines are shown between the results of both models for the same language for visual clarity. We note that both models had little to no coverage of the scripts for \texttt{ml} and \texttt{gu}.}
    \label{fig:script-analysis}
\end{figure*}

\paragraph{Tokenization Analysis.} The languages we study in Camellia span a wide range of writing systems. Chinese is written using a logographic script. Japanese combines logographic characters (Kanji) with two syllabaries (Hiragana and Katakana). Korean uses Hangul, an alphabet arranged into block-like characters. In contrast, the remaining languages use alphabetic systems, including Perso-Arabic script for Urdu, Brahmic scripts for Indian languages. The way these langauges are tokenized varies from one model to another.

To study the impact of tokenization differences across different models, we analyze the relationship between model performance and the coverage of each language’s script within the tokenizer vocabulary. Specifically, for each model, we compute the percentage of tokens in its vocabulary containing at least one character from the script. We identify script-specific characters using their Unicode ranges (e.g., \texttt{\textbackslash u4E00}-\texttt{\textbackslash u9FFF} for Chinese, \texttt{\textbackslash u1100}–\texttt{\textbackslash u11FF} for Hangul, etc.). For Vietnamese, which uses the Latin alphabet with diacritical marks, we specifically count tokens containing such Vietnamese-specific markers (e.g., \texttt{ă}, \texttt{â}, \texttt{ê}, \texttt{ô},   \texttt{à}, \texttt{á}, \texttt{ả}, \texttt{ã}, etc.), ensuring we reflect tokens containing Vietnamese-specific characters rather than generic Latin script.

Figure~\ref{fig:script-analysis} presents the CBS results for Llama-3.3-70B and Qwen-2.5-72B on all languages, plotted against each model’s tokenizer script coverage. Overall, we observe that higher script coverage in a tokenizer tends to yield better performance (i.e., lower CBS). This trend is especially clear for Chinese, Japanese, and Korean, where Qwen outperforms Llama, consistent with Qwen’s stronger coverage of these scripts. In contrast, for Hindi, Marathi, Urdu, and Vietnamese, the pattern reverses: Llama performs better, reflecting its better coverage of the scripts of these languages. As noted in prior studies \cite{foroutan2025parity}, tokenization algorithms such as BBPE are trained on corpora with imbalanced language and script representation, which can place languages with underrepresented scripts at a disadvantage.

\subsection{Sentiment Association}

\paragraph{Test Set Sizes.} Table~\ref{tab:test-set-size-sentiment} reports the exact size of the test sets used in our sentiment association experiment (\S~\ref{subsec:sentiment-association}). The test set of each language is constructed by taking each masked context in \texttt{Camellia-Grounded} and \texttt{Camellia-Neutral} which are annotated for sentiment and creating 50 samples out of each context by replacing the \texttt{[MASK]} by 50 randomly sampled entities associated with the respective Asian culture or Western culture. Thus, the size of the Asian and Western test sets for each language is the same. We obtain test sets that range from generally range from 13,000 to 24,000 samples, depending on the amount of masked contexts we obtained in each language during data collection. We note that for Urdu the size of the test sets are smaller (2,550 samples each for Pakistani and Western) due to the language’s low-resource nature and the limited availability of masked contexts.

\begin{figure*}[t]
    \centering
    \includegraphics[width=\linewidth]{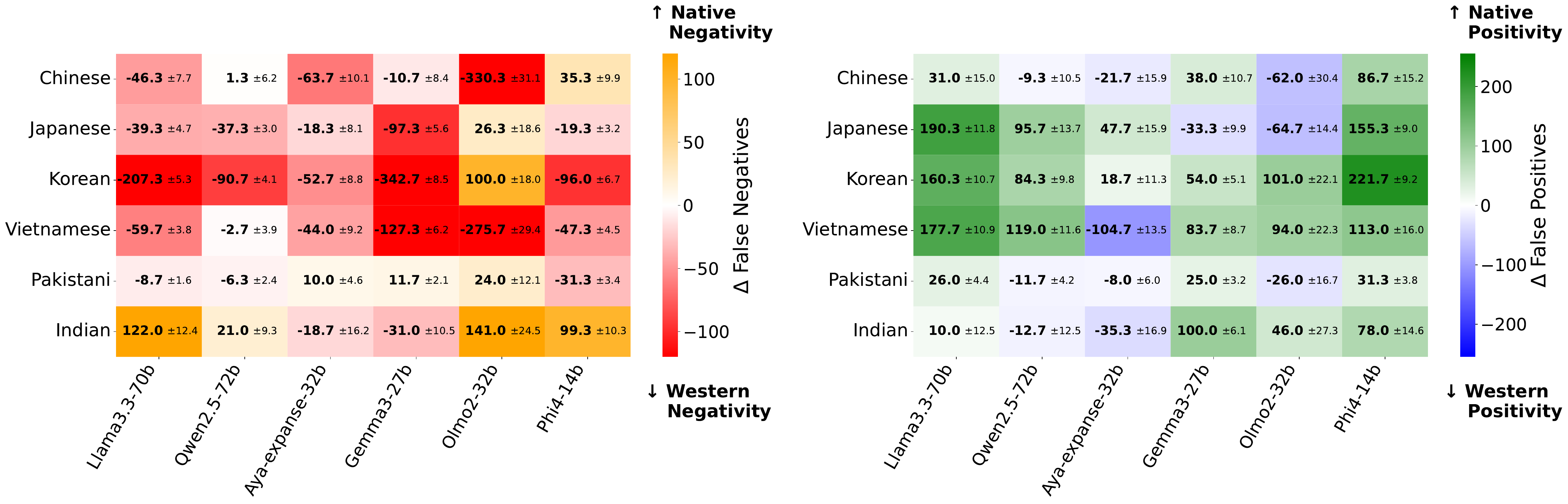}
    \caption{Differences in False Negative (FN) and False Positive (FP) sentiment predictions by LLMs on \texttt{Camellia} contexts filled with Asian vs Western entities, when tested in English. Results are averaged across 3 runs of 50 randomly sampled Asian vs Western entities in each culture.}
    \label{fig:sentiment-results-english}
\end{figure*}

\paragraph{Results when testing in English.} Figure~\ref{fig:sentiment-results-english} shows the result of our sentiment association experiment when testing LLMs on the parallel English translations of the entities and contexts in each culture. In certain cases, the behavior of some models such as Gemma in English is consistent to when we tested in Asian languages, with generally more Western negativity and more positivity towards native Asian entities of each culture. There are certain cases where trends from the same model become different, such as for the Llama model, where it becomes more positive with native Asian entities in English.

\begin{table}[t]
\centering
\begin{tabular}{@{}lc@{}}
\toprule
\textbf{Language} & \textbf{Test Set Size} \\ \midrule
\texttt{zh} & 17,900 \\
\texttt{ja} & 13,850 \\
\texttt{ko} & 24,500 \\
\texttt{vi} & 17,550 \\
\texttt{ur} & 2,550 \\
\texttt{hi} & 19,882 \\
\texttt{ml} & 19,882 \\
\texttt{mr} & 19,882 \\
\texttt{gu} & 19,882 \\ \bottomrule
\end{tabular}
\caption{Size of the native Asian and Western test sets used in our sentiment association experiment for each language.}
\label{tab:test-set-size-sentiment}
\end{table}

\subsection{Extractive QA}
\label{appendix:extractive-qa-results}

\paragraph{Test Set Sizes.} Table~\ref{tab:test-set-size-qa} reports the exact size of the test sets used in our entity extractive QA experiment (\S~\ref{subsec:extractive-qa}). The test set of each language is constructed by taking each masked context in \texttt{Camellia-QA} and creating 50 samples out of each context by replacing the \texttt{[MASK]} by 50 randomly sampled entities associated with the respective Asian culture or Western culture.

\begin{table}[h]
\centering
\begin{tabular}{@{}lc@{}}
\toprule
\textbf{Language} & \textbf{Test Set Size} \\ \midrule
\texttt{zh} & 3,200 \\
\texttt{ja} & 3,000 \\
\texttt{ko} & 3,500 \\
\texttt{vi} & 3,900 \\
\texttt{ur} & 2,900 \\
\texttt{hi} & 2,350 \\
\texttt{ml} & 2,350 \\
\texttt{mr} & 2,350 \\
\texttt{gu} & 2,350 \\ \bottomrule
\end{tabular}
\caption{Size of the native Asian and Western test sets used in our extractive QA experiment.}
\label{tab:test-set-size-qa}
\end{table}

\paragraph{Detailed Extractive QA Results.} Tables~\ref{tab:qa-en-1} and \ref{tab:qa-en-2} show the detailed accuracy results on the extractive QA task. We compute accuracy based on the exact match of identifying the entity in the context. We observe large accuracy gaps between sets containing Asian and Western entities when testing in the respective Asian language of each culture, where LLMs mostly perform better at extracting Asian-associated entities. In contrast, these gaps are negligible in English in nearly all cases (2\%-5\% gaps). In a couple of cases, large gaps in English are observed (Pakistani vs Western entities in Llama and Qwen, Indian vs Western entities in Qwen). 

\begin{table*}[h]
\centering
\begin{adjustbox}{width=\linewidth}
\begin{tabular}{@{}lcccccc|cccccc@{}}
\cmidrule(l){2-13}
 & \multicolumn{6}{c|}{\textbf{Llama3.3-70b}} & \multicolumn{6}{c}{\textbf{Qwen2.5-72b}} \\ \cmidrule(l){2-13} 
\multicolumn{1}{r}{\small{Test Lang}} & \multicolumn{3}{c}{\textbf{Respective Asian}} & \multicolumn{3}{c|}{\textbf{English}} & \multicolumn{3}{c}{\textbf{Respective Asian}} & \multicolumn{3}{c}{\textbf{English}} \\ \cmidrule(l){2-13} 
\multicolumn{1}{r}{\small{Culture}} & \textit{Asian} & \textit{Western} & \multicolumn{1}{c|}{$\Delta$Acc} & \textit{Asian} & \textit{Western} & $\Delta$Acc & \textit{Asian} & \textit{Western} & \multicolumn{1}{c|}{$\Delta$Acc} & \textit{Asian} & \textit{Western} & $\Delta$Acc \\ \midrule
Chinese & 94.81 & 96.13 & \cellcolor[HTML]{DCF3F4}-1.32 & 91.42 & 91.11 & \cellcolor[HTML]{FFF3EA}0.30 & 95.46 & 95.03 & \cellcolor[HTML]{DCF3F4}0.43 & 88.57 & 91.41 & \cellcolor[HTML]{FFF3EA}-2.84 \\
Japanese & 91.49 & 83.94 & \cellcolor[HTML]{98DBE0}7.55 & 92.48 & 89.77 & \cellcolor[HTML]{FFF3EA}2.72 & 88.47 & 69.60 & \cellcolor[HTML]{46BDC6}18.87 & 88.44 & 83.90 & \cellcolor[HTML]{FFF3EA}4.53 \\
Korean & 91.74 & 82.06 & \cellcolor[HTML]{98DBE0}9.69 & 92.34 & 91.69 & \cellcolor[HTML]{FFF3EA}0.66 & 91.17 & 74.70 & \cellcolor[HTML]{46BDC6}16.47 & 85.14 & 87.63 & \cellcolor[HTML]{FFF3EA}-2.49 \\
Vietnamese & 74.78 & 88.31 & \cellcolor[HTML]{46BDC6}-13.53 & 91.70 & 89.75 & \cellcolor[HTML]{FFF3EA}1.95 & 73.67 & 88.00 & \cellcolor[HTML]{46BDC6}-14.33 & 83.44 & 87.05 & \cellcolor[HTML]{FFF3EA}-3.61 \\
Pakistani & 75.42 & 80.13 & \cellcolor[HTML]{DCF3F4}-4.71 & 99.66 & 89.11 & \cellcolor[HTML]{FFC092}10.54 & 67.73 & 72.71 & \cellcolor[HTML]{DCF3F4}-4.99 & 99.77 & 87.61 & \cellcolor[HTML]{FFC092}12.16 \\
Indian \texttt{(hi)} & 95.45 & 85.40 & \cellcolor[HTML]{46BDC6}10.05 & 98.31 & 91.59 & \cellcolor[HTML]{FFDDC4}6.71 & 70.38 & 66.74 & \cellcolor[HTML]{DCF3F4}3.63 & 98.06 & 87.38 & \cellcolor[HTML]{FFC092}10.67 \\
Indian \texttt{(ml)} & 76.09 & 62.94 & \cellcolor[HTML]{46BDC6}13.15 & --- & --- & --- & 55.73 & 51.51 & \cellcolor[HTML]{DCF3F4}4.22 & --- & --- & --- \\
Indian \texttt{(mr)} & 94.45 & 83.38 & \cellcolor[HTML]{46BDC6}11.07 & --- & --- & --- & 48.58 & 46.90 & \cellcolor[HTML]{DCF3F4}1.68 & --- & --- & --- \\
Indian \texttt{(gu)} & 87.56 & 73.12 & \cellcolor[HTML]{46BDC6}14.44 & --- & --- & --- & 50.43 & 44.40 & \cellcolor[HTML]{80D2D8}6.02 & --- & --- & --- \\ \bottomrule
\end{tabular}
\end{adjustbox}
\caption{Detailed accuracy results for Llama3.3-70b and Qwen2.5-72b on the extractive QA task when tested in the respective Asian language of each culture vs. in English. }
\label{tab:qa-en-1}
\end{table*}

\begin{table*}[h]
\centering
\begin{adjustbox}{width=\linewidth}
\begin{tabular}{@{}lcccccccccccc@{}}
\cmidrule(l){2-13}
 & \multicolumn{6}{c}{\textbf{Aya-expanse-32b}} & \multicolumn{6}{c}{\textbf{Gemma3-27b}} \\ \cmidrule(l){2-13} 
\multicolumn{1}{r}{\small{Test Lang}} & \multicolumn{3}{c}{\textbf{Respective Asian}} & \multicolumn{3}{c|}{\textbf{English}} & \multicolumn{3}{c}{\textbf{Respective Asian}} & \multicolumn{3}{c}{\textbf{English}} \\ \cmidrule(l){2-13} 
\multicolumn{1}{r}{\small{Culture}} & \textit{Asian} & \textit{Western} & \multicolumn{1}{c|}{$\Delta$Acc} & \textit{Asian} & \textit{Western} & \multicolumn{1}{c|}{$\Delta$Acc} & \textit{Asian} & \textit{Western} & $\Delta$Acc & \textit{Asian} & \textit{Western} & $\Delta$Acc \\ \midrule
Chinese & 87.08 & 84.24 & \cellcolor[HTML]{DCF3F4}2.84 & 81.08 & 86.91 & \multicolumn{1}{c|}{\cellcolor[HTML]{FFF3EA}-5.83} & 91.58 & 92.94 & \cellcolor[HTML]{DCF3F4}-1.36 & 84.13 & 89.76 & \cellcolor[HTML]{FFF3EA}-5.63 \\
Japanese & 86.96 & 78.12 & \cellcolor[HTML]{80D2D8}8.84 & 83.77 & 84.51 & \multicolumn{1}{c|}{\cellcolor[HTML]{FFF3EA}-0.73} & 81.84 & 65.44 & \cellcolor[HTML]{46BDC6}16.40 & 83.97 & 87.19 & \cellcolor[HTML]{FFF3EA}-3.22 \\
Korean & 93.20 & 79.26 & \cellcolor[HTML]{46BDC6}13.94 & 95.51 & 94.09 & \multicolumn{1}{c|}{\cellcolor[HTML]{FFF3EA}1.43} & 92.43 & 84.49 & \cellcolor[HTML]{80D2D8}7.94 & 96.71 & 94.17 & \cellcolor[HTML]{FFF3EA}2.54 \\
Vietnamese & 76.56 & 73.73 & \cellcolor[HTML]{DCF3F4}2.83 & 91.09 & 92.97 & \multicolumn{1}{c|}{\cellcolor[HTML]{FFF3EA}-1.88} & 93.87 & 89.72 & \cellcolor[HTML]{DCF3F4}4.15 & 97.66 & 96.01 & \cellcolor[HTML]{FFF3EA}1.65 \\
Pakistani & 66.66 & 66.53 & \cellcolor[HTML]{DCF3F4}0.12 & 97.61 & 93.08 & \multicolumn{1}{c|}{\cellcolor[HTML]{FFF3EA}4.54} & 81.75 & 60.64 & \cellcolor[HTML]{46BDC6}21.11 & 99.53 & 95.00 & \cellcolor[HTML]{FFF3EA}4.54 \\
Indian \texttt{(hi)} & 86.39 & 74.85 & \cellcolor[HTML]{46BDC6}11.54 & 94.62 & 93.55 & \multicolumn{1}{c|}{\cellcolor[HTML]{FFF3EA}1.07} & 85.72 & 78.91 & \cellcolor[HTML]{80D2D8}6.81 & 98.52 & 95.26 & \cellcolor[HTML]{FFF3EA}3.25 \\
Indian \texttt{(ml)} & 70.46 & 59.52 & \cellcolor[HTML]{46BDC6}10.93 & --- & --- & \multicolumn{1}{c|}{---} & 52.87 & 43.85 & \cellcolor[HTML]{80D2D8}9.01 & --- & --- & --- \\
Indian \texttt{(mr)} & 81.84 & 69.20 & \cellcolor[HTML]{46BDC6}12.64 & --- & --- & \multicolumn{1}{c|}{---} & 86.80 & 83.29 & \cellcolor[HTML]{DCF3F4}3.50 & --- & --- & --- \\
Indian \texttt{(gu)} & 65.19 & 52.30 & \cellcolor[HTML]{46BDC6}12.89 & --- & --- & \multicolumn{1}{c|}{---} & 86.30 & 79.76 & \cellcolor[HTML]{80D2D8}6.54 & --- & --- & --- \\ \bottomrule
\end{tabular}
\end{adjustbox}
\caption{Detailed accuracy results for Aya-expanse-32b and Gemma3-27b on the extractive QA task when tested in the respective Asian language of each culture vs. in English. }
\label{tab:qa-en-2}
\end{table*}

\end{document}